\documentclass[aoas]{imsart}

\RequirePackage{amsthm,amsmath,amsfonts,amssymb}
\RequirePackage[authoryear]{natbib}
\RequirePackage[colorlinks,citecolor=blue,urlcolor=blue]{hyperref}
\RequirePackage{graphicx}
\usepackage{algorithm}
\usepackage{algorithmicx}
\usepackage{array}
\usepackage{tikz}
\usetikzlibrary{matrix,decorations.pathreplacing}

\startlocaldefs
\theoremstyle{plain}

\theoremstyle{remark}

\numberwithin{equation}{section}
\theoremstyle{plain}

\newcommand{\real}{\ensuremath{\mathbb{R}}}

\newtheorem{lemma}{Lemma}

\newtheorem{remark}{Remark}

\DeclareMathOperator{\Tr}{Tr}
\DeclareMathOperator*{\argmax}{arg\,max}
\DeclareMathOperator*{\argmin}{arg\,min}
\DeclareMathOperator{\vect}{vec}

\endlocaldefs

\begin{document}

\begin{frontmatter}
\title{Statistical shape analysis of brain arterial networks (BAN)}

\begin{aug}
\author[A]{\fnms{Xiaoyang} \snm{Guo}\ead[label=e1]{xiaoyang.guo.fl@gmail.com}},
\author[A]{\fnms{Aditi} \snm{Basu Bal}\ead[label=e2]{ab18z@my.fsu.edu}}
\and
\author[B]{\fnms{Tom} \snm{Needham}\ead[label=e3]{tneedham@fsu.edu}}
\author[A]{\fnms{Anuj} \snm{Srivastava}\ead[label=e4]{anuj@stat.fsu.edu}}

\address[A]{Department of Statistics,
Florida State University,
\printead{e1,e2,e4}}

\address[B]{Department of Mathematics,
Florida State University,
\printead{e3}}
\end{aug}

\begin{abstract}
Structures of arterial networks in the human brain, termed Brain Arterial Networks or BANs -- that are complex arrangements of individual arteries, their 
branching patterns, and 
inter-connectivities -- play an essential role in characterizing and understanding brain physiology. 
One would like tools for statistically analyzing the shapes of BANs, i.e., quantifying shape differences, comparing a population of subjects, and studying the effects of covariates on these shapes. 
This paper mathematically represents and statistically analyzes 
BAN shapes as {\it elastic shape graphs}. Each elastic shape graph consists of nodes, or points in 3D, connected by some 3D curves, or edges, with arbitrary shapes.
We develop a mathematical representation,  
a Riemannian metric and other geometrical tools, such as computations of geodesics, means, covariances, and PCA, for helping analyze BANs as elastic graphs. We apply this analysis to BANs after dividing them into four components -- top, bottom, left, and right. 
The framework is then used to generate shape summaries of BANs from 92 subjects and study the effects of age and gender on shapes of BAN components. We conclude that while gender effects require further
investigation, age has a clear, quantifiable effect on BAN shapes. Specifically, we find an increased variance in BAN shapes as age increases. 
\end{abstract}


\end{frontmatter}

\section{Introduction}

The human brain is one of the most sophisticated organs in the human body and serves as the center of the nervous system. It requires a significant amount of energy to accomplish its designated tasks and  
a complex network of arteries is used to supply the necessary oxygen and nutrition to parts of the brain.
This network, called the {\it Brain arterial network} or BAN, is central
to maintaining standard anatomical functionality of the brain. 
The structure or morphology of BANs determines their effectiveness in providing supply lines and in
characterizing and diagnosing brain health.
Consequently, a statistical analysis of BANs, 
which constitutes representing and analyzing shape variability within and 
across human populations is a significant problem area.
However, this analysis is challenging because
the BANs have complicated structures, with tremendous variability in terms of 
branching, winding, and merging nature of arteries on the one hand, and the shapes and sizes of 
arteries on the other.

\begin{figure}
\begin{center}
\begin{tabular}{m{3cm}|m{4cm}m{4cm}}
\includegraphics[width=0.15\textwidth]{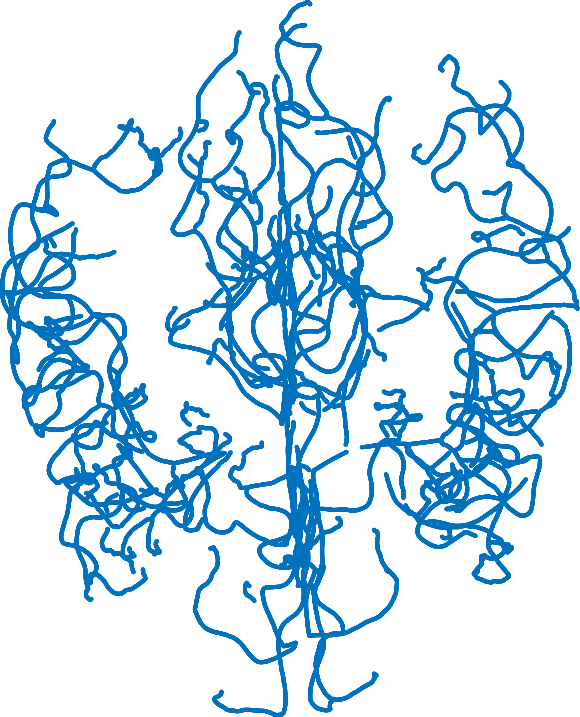}&
\includegraphics[width=0.3\textwidth]{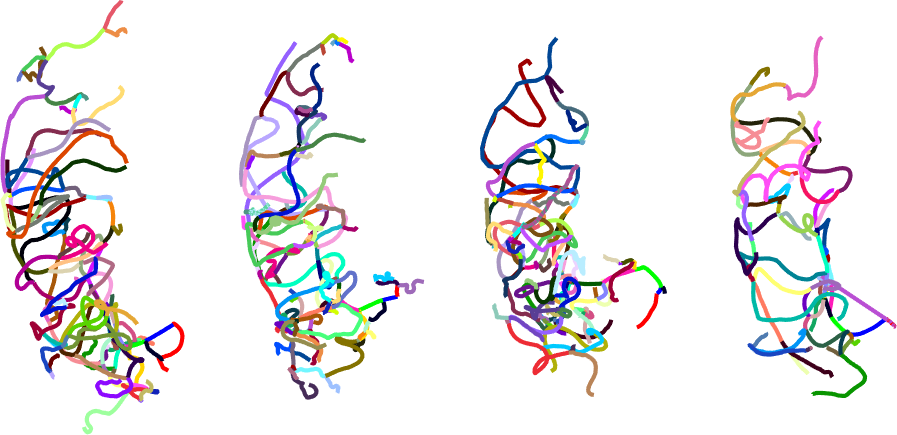}
&\includegraphics[width=0.3\textwidth]{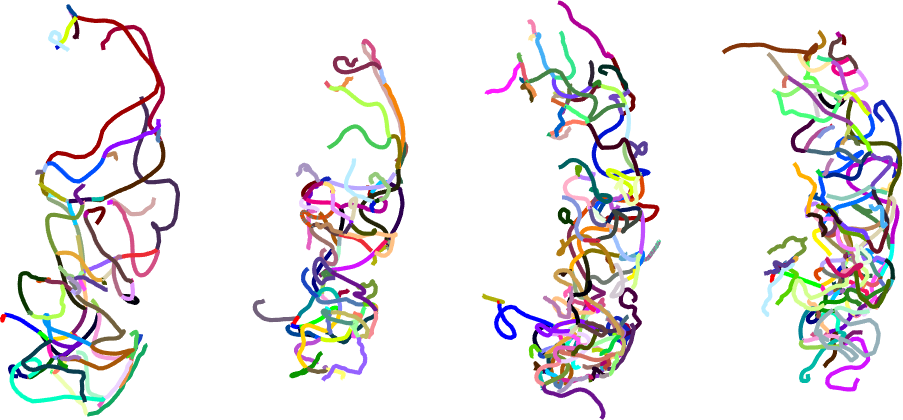}\\
\scriptsize{Dense Point Cloud} & Left & Right \\
\includegraphics[width=0.15\textwidth]{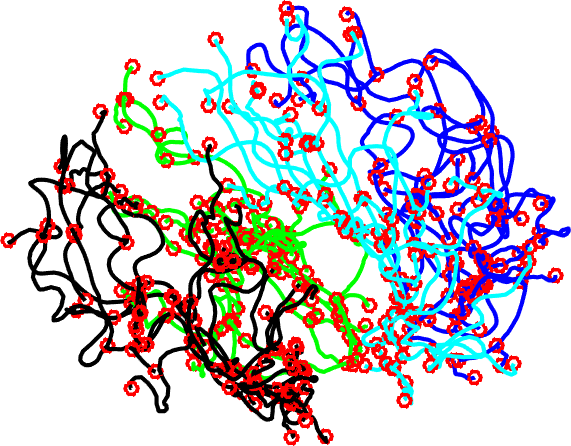}&
\includegraphics[width=0.3\textwidth]{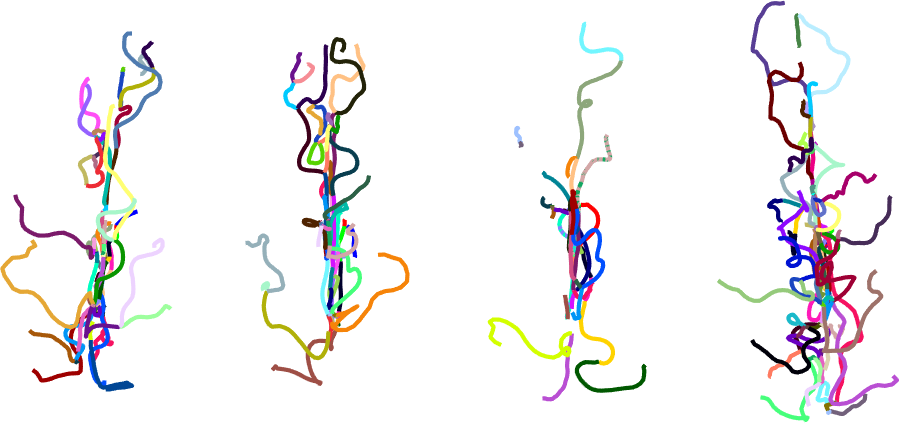}
&\includegraphics[width=0.31\textwidth]{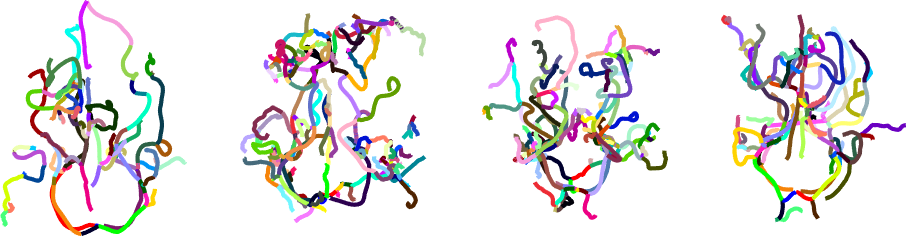} \\
\scriptsize{Network of 3D Curves} & Top & Bottom \\
\end{tabular}
\end{center}
\caption{Examples of brain arterial networks. Left side is the different representations of BAN data. 
Cyan, black, green, and blue display the four components: 
top, left, bottom, and right, respectively. Red circles denote nodes. Right side is examples of four components of BANs from different subjects.}
\label{fig:example}
\end{figure}

Fig. \ref{fig:example} shows some examples of BANs for different human subjects. 
Each BAN is reconstructed from a 3D Magnetic Resonance Angiography (MRA) image using a tube-tracking vessel segmentation algorithm~[\cite{aydin2009principal,aylward2002initialization}].
Given the complex nature of BANs, previous analyses have primarily focused
on first extracting some low-dimensional features from the original data, followed by a statistical analysis of these features. 
One of the earliest analyses~[\cite{bullitt2010effects}] focused on some simple geometrical summaries, 
such as the numbers and the lengths of the arteries. 
In recent years, the extracted features have become more sophisticated. 
For instance, \cite{bendich2016persistent} uses tools from Topological Data Analysis (TDA), where one
extracts certain mathematical features (e.g., {\it persistent homology}) from the data and compares these
features using certain metrics~[\cite{wasserman2018topological}]. 
However, the difficulty in such {\it feature-based} approaches is that these representations
are typically not invertible. Feature extraction usually represents only partial information about the original objects, 
forming a many-to-one mapping
(from the object space to a feature space). Moreover, it is not clear as to which set of shapes share the same topological features. Because of this lack of invertibility, it is not easy to
map solutions or statistical inferences back to the object space.
This paper takes a more ambitious approach, where we develop a statistical analysis of BANs
in the original space itself without resorting to extracting any features. In the process, we seek solutions -- shape summaries, shape PCA, and shape models -- that can be studied as BANs themselves.

Besides BANs, there are other biological and anatomical objects with similar shape architectures, displaying complex filamentary shapes.
Examples include retinal blood vessels~[\cite{hoover2000locating}],  vein structures in fruit fly wings~[\cite{sonnenschein2015image}] and  neurons~[\cite{kong2005diversity}]. A defining characteristic of these objects is that they
are composed of a network of 2D/3D curves, each with arbitrary shapes and sizes. Moreover, these curves merge and branch at arbitrary junctions and result in intricate patterns of pathways. 
While we mainly focus on BANs in this paper, the proposed framework is also applicable to these other application domains.
Statistical shape analysis of such objects is difficult because, to quantify shape differences, one needs to consider the numbers, locations, branchings, and shapes of individual curves. In particular, 
one has to solve a difficult problem of {\it registration of points and parts} 
across objects, {\it i.e.,} which points on a branch on one object matches
with which points or parts on the other. 

The field of shape analysis has steadily gained relevance and activity over the last two decades.
This rise is fueled by the availability of multimodal, high-dimensional data that records objects of interest in various contexts and applications. Shapes of objects help characterize their identity, classes, movements, and roles in larger scenes. 
Consequently, many approaches have been developed for comparing, summarizing, modeling, testing, and tracking shapes in a static image or video data. 
Statistical shape analysis requires mathematical representations and proper metrics.
While early methods generally relied on discrete representations of shapes 
~[\cite{kendall1984shape,kendall-barden-carne,small-shapes,dryden2016statistical}], more
recent methods have focused on continuous objects such as scalar functions~[\cite{srivastava2016functional}], 
Euclidean curves~[\cite{younes-distance2,klassen2004analysis,younes-michor-mumford-shah:08,srivastava2016functional}], 
and 3D surfaces~[\cite{jermyn2017elastic}]. The main motivation for this paradigm shift comes from the need to address the {\it registration} problem, 
considered the most challenging issue in shape analysis. 
As mentioned earlier, registration refers to establishing a correspondence between points or features across objects and is an important
ingredient in comparing shapes.
Continuous representations of objects use actions of the re-parameterization groups to help solve dense registration problems. Furthermore, they use elastic Riemannian metrics -- which are invariant to the actions of re-parameterization groups -- and some simplifying square-root representations to develop 
very efficient techniques for comparing and analyzing shapes. 

While elastic Riemannian shape analysis is considered well developed for elementary objects -- Euclidean 
curves~[\cite{younes-michor-mumford-shah:08,srivastava2010shape}], 
manifold-valued curves~[\cite{zhang2018phase,lebrigant-2019}], 3D surfaces~[\cite{jermyn2012elastic}],  trees~[\cite{duncan2018statistical,wang2020statistical}] - the problem of 
analyzing more complex
objects remains less explored. In other words, the past work has mainly focused on objects that exhibit only the geometrical variabilities while being of the same or similar topologies. Similar topologies help 
pose the registration problem as that of optimal (diffeomorphic) re-parameterization of the common domain. 
In this paper, we are concerned with comparing BANs that can potentially differ in both
topologies and geometries. 
 \\
 
\noindent{\bf Specific Goals}: Our goal here is to develop a suite of techniques for statistical analysis of BAN
shapes. Specifically, we seek: (1) a {\bf shape metric} that is invariant to the
usual shape-preserving transformations, (2)  elastic {\bf registration} of parts across BANs, (3) computation
of {\bf geodesic} paths between any two BANs and (4) computation of {\bf statistical summaries} -- mean, covariance, PCA -- in the shape space of BANs. These tools, in turn, can be used for analysis, clustering, classification,  
and modeling of shapes in conjunction with other machine learning methods. We reiterate that existing techniques may provide some but not all of these solutions.  

Our approach is to view BANs as extensions of traditional graphs with the usual node-edge representations. 
The difference lies in that edge characterization is now more sophisticated -  each edge is a full shape of a curve connecting the corresponding nodes, 
as shown on the right side of Fig.~\ref{fig:example}. These graphs, called {\it elastic graphs}, are
represented by their adjacency matrices, with matrix elements given by the shapes of the corresponding edges. Since the ordering of nodes in these graphs and the subsequent indexing of adjacency matrices is arbitrary we model this variability using the action of the permutation group and represent each shape as an orbit under this group. Then, we develop techniques for optimization under this permutation group (also known as {\it graph matching}), 
leading to computations of geodesics and summaries under the induced
metric on the Riemannian quotient space, termed the {\it graph shape space}. 
There is no current geometrical framework for the shape analysis of such graphical objects. While TDA and other such methods can measure dissimilarity in shapes, this paper provides more comprehensive statistical quantities, such as mean, covariance, and principal components, for a more in-depth shape
summarization and modeling. 

In order to further tame the complexity and to improve registration results, we divide a BAN into four components -- top, bottom, left, and right -- as shown in different colors on the left side of Fig.~\ref{fig:example}.
This division is not always precise, but the subsequent statistical analysis is generally
robust to small variations in the division.
Then, we study each of these components independently, as separate objects. Right side of Fig.~\ref{fig:example} shows some 
examples of these four components from different subjects.

\section{Proposed Mathematical Framework}
\label{sec:framework}

We now present a mathematical framework for representing elastic graphs, keeping in mind that
BAN components form the focus of this study. 
The proposed framework can be viewed as an extension of some previous works on graphs
~[\cite{calissano2020populations,Chowdhury_2020_CVPR_Workshops,guo2021quotient,jain2009structure,jain2011graph,jain2012learning}]. 
However, those past works are restricted to only scalar-weighted graphs and do not consider more sophisticated
features such as edge shapes. 

\subsection{Elastic Graph Representation}
We are interested in objects made of several 3D curves, with arbitrary shapes and placements, that merge and branch at arbitrary junctions and result in networks of pathways. 
We will represent them as graphs with nodes corresponding to junctions and edges corresponding to the shapes of 3D curves connecting the nodes. Here we assume that any two nodes
are connected directly by at most one curve. 
An edge attributed graph $G$ is an ordered pair $(V,a)$, 
where $V$ is a set of nodes and $a$ is an edge attribute function: $a: V \times V \rightarrow {\cal S}$.
${\cal S}$ is the shape space of elastic 3D curves that is briefly discussed next.

The edges in elastic graphs are Euclidean curves and to analyze their shapes 
we use elastic shape analysis framework described in~[\cite{srivastava2016functional}].
Let $\beta: [0,1] \rightarrow \mathbb{R}^n, n=2$ or $3$, be an absolutely continuous function, 
representing a parametrized curve.
Define the square root velocity function (SRVF) of $\beta$ as:
$q(t)= \frac{\dot{\beta}(t)}{\sqrt{|\dot{\beta}(t)|}} \in \real^n$, if $|\dot{\beta}(t)| \neq 0$ and zero otherwise. 
One can recover $\beta$ from its SRVF using $\beta(t) = \beta(0) + \int_{0}^t q(s)|q(s)| ds$.
If $\beta$ is absolutely continuous, the SRVF is square-integrable, {\it i.e.}, $q \in \mathbb{L}^2([0,1],\real^n)$
or simply $\mathbb{L}^2$.
It can be shown that the $\mathbb{L}^2$ norm on SRVF space is an elastic Riemannian metric on the original curve space.
Therefore, one can compute the elastic (i.e., rotation and reparameterization-invariant) distance between two curves $\beta_1, \beta_2$ using  
$d(\beta_1,\beta_2) = \|q_1-q_2 \|_{\mathbb{L}^2}$.
One of the most important challenges in shape analysis is registration, a dense correspondences across curves. 
Let $\gamma : [0,1] \rightarrow [0,1]$ be a diffeomorphism; in classical elastic shape analysis, diffeomorphisms are restricted to be orientation-preserving ($\gamma(0) = 0$ and $\gamma(1) = 1$), but in the proposed framework we make no such restriction. The action of the diffeomorphism group on an SRVF $q$ is $q \ast \gamma = (q \circ \gamma)\sqrt{\dot{\gamma}}$.
This is same the expression as the SRVF of the re-parameterized curve: $\beta \circ \gamma$.
To register points across curves, one mods out this re-parametrization group as follows.
Each shape can be represented by orbits under the
re-parametrization group: $[q] = \{q\ast \gamma | \gamma \in \Gamma\}$. 
The set of all orbits is the shape space of curves in $\real^n$ is denoted $\mathcal{S} = \{[q] | q \in \mathbb{L}^2\}$.
The shape metric is then given by: 
$d_s([q_1],[q_2]) = \inf_{\gamma} \|q_1 - (q_2\ast \gamma)\|$. 
By including both orientation-preserving and reversing diffeomorphisms, the shape distance
between a curve $(\beta(t))$  and its parameterized reflection  ($\beta(1-t)$) is zero.

One can use this metric $d_s$ to define and compute
averages of shapes of curves and their PCA analysis. 
For further details, we refer the reader to the textbook~[\cite{srivastava2016functional}].

\begin{remark}
Note that traditionally one further removes the rotation and scale of curves from considerations in 
shape analysis, but here these two variables are integral to the shapes of arteries and the network as a whole.
So, we do not remove them. We do perform a global scale and rotational alignment of the whole BAN when comparing
it with another BAN. 
\end{remark}

\begin{figure}
\centering
\includegraphics[width=0.7\textwidth]{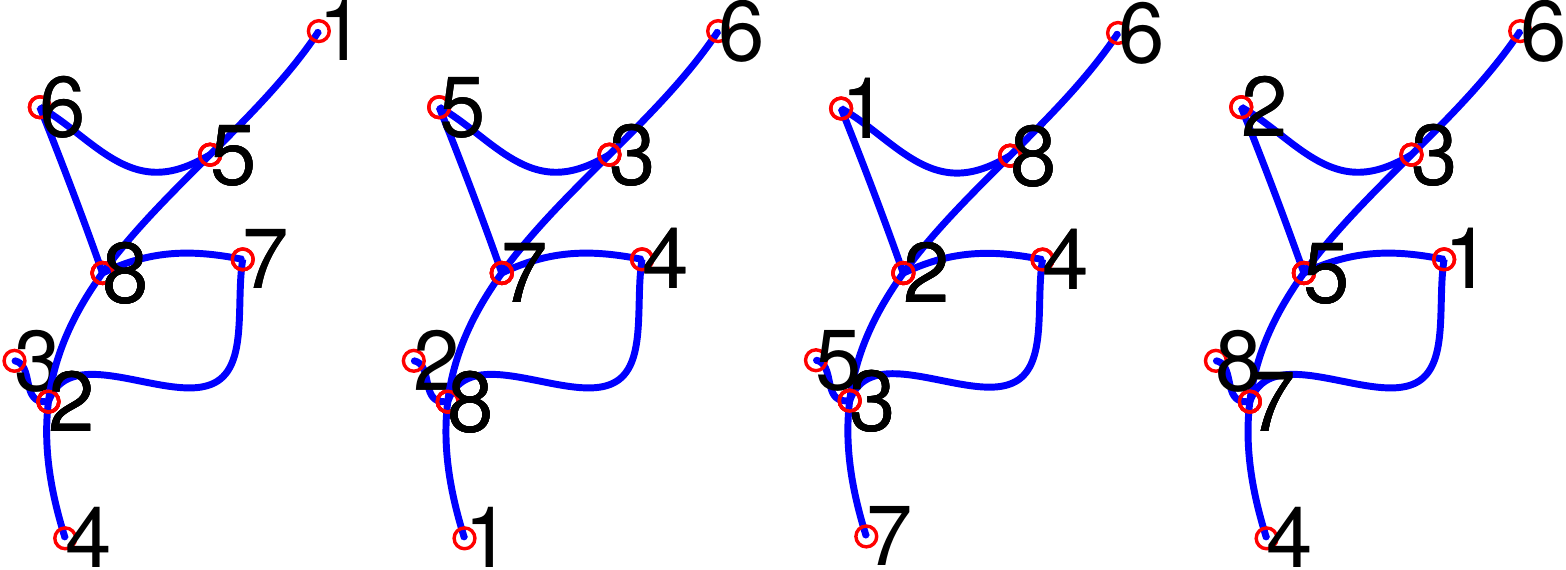}
\caption{Four graphs with the same shape but with different node labels.}
\label{fig:node_order}
\end{figure}

Returning to the elastic graph, 
the shape $a(v_i, v_j)=[q_{ij}]$ characterizes the shape of curve connecting the nodes $v_i, v_j \in V, i \neq j$.
Since the representation $[q_{ij}]$ includes both a curve and its parameterized reflection, this edge attribute is 
directionless. 
Assuming that the number of nodes is  $n$, $G$ can be represented by its adjacency matrix 
$A = \{a_{ij}\} \in {\cal S}^{n \times n}$, where the element $a_{ij} = a(v_i, v_j)$.
For an undirected graph $G$, we have $a(v_i, v_j) = a(v_j, v_i)$ and therefore $A$ is a symmetric matrix. 
As an example,  the adjacency matrix of the third graph shown in Fig. \ref{fig:node_order} is given by
\[
\tiny \begin{pmatrix} 
\mathbf{0} & [q_{12}]  & \mathbf{0} & \cdots & [q_{18}]   \\
[q_{21}]      & \mathbf{0} & [q_{23}]      & \cdots & [q_{28}]\\
\mathbf{0} & [q_{32}]      & \mathbf{0} & \cdots & \mathbf{0} \\
\vdots       & \vdots        & \vdots        & \ddots & \vdots \\
 [q_{81}] & [q_{82}]     & \mathbf{0}  & \cdots & \mathbf{0}\\
\end{pmatrix} \in {\cal S}^{n \times n},
\]
where $\mathbf{0}$ denotes a null edge. It implies that the corresponding nodes are not connected and 
we substitute the constant zero function ${\bf  0} \in \mathbb{L}^2$ as its shape. 
Note that $[{\bf 0}] = {\bf 0}$ is an element of the shape space ${\cal S}$
and one can perform all the standard operations, such as computing shape distance, computing
averages, or performing tangent PCA, with null edges using the geometry of ${\cal S}$. 
A little later we will also use the concept of null nodes, extraneous nodes
that are attached to the existing graphs using null edges. 
This use of null nodes facilitates improved matching and comparisons of graphs.
As mentioned earlier, we will assume that there are no self-loops in BANs and
therefore the diagonal entries in $A$ will also be null. 
The set of all such adjacency matrices is given by ${\cal A} = \{ A \in {\cal S}^{n \times n} | A = A^T, \text{diag}(A) = \mathbf{0}\}$.

In some situations, it may be useful to include the node attributes
also in graph comparisons.
In this paper, we use the degree of a node, {\it i.e.}, the number of other nodes connected to a node by non-trivial edges, 
as its attribute. Let $u \in \real^n$ denote the vector of node attributes associated with the graph. Note that the ordering of nodes in $u$ matches the ordering of nodes in $A$.
Combining the node-edge representations we get a joint representation $B = (A, u) \in {\cal A} \times \real^n \equiv {\cal B}$.

Next we define a metric for comparing graphs. We will start by assuming that all the graphs have the same number of nodes, but will 
include graphs with arbitrary number of nodes shortly.
For any two $A_1, A_2 \in {\cal A}$, 
with the corresponding entries $a_{ij}^1$ and $a_{ij}^2$, respectively, we define the edge metric 
$d_a$ to be:
$d_a(A_1, A_2) \equiv
\sqrt{\sum_{i,j} d_s(a_{ij}^1, a_{ij}^2)^2}$.
$d_a$ quantifies the differences between adjacencies of graphs, $A_1$ and $A_2$, where
$d_s$ is the shape metric for curves as defined above. 
Additionally, let the matrix of distances between the node
attributes $u_1, u_2$ to be: $D = | u_1 {\bf 1}_n^T - {\bf 1}_n u_2^T|$, 
where ${\bf 1}_n$ is an $n$-vector of  all ones. 
Thus, $D_{ij}$ is difference between the attributes of nodes $v_i^1$ and $v_j^2$.
This leads to a node metric between two graphs according to $d_v=\Tr(D)$. 
Finally, we can combine the two metrics to reach a composite metric:  $d_b(B_1, B_2) \equiv d_a + \lambda \Tr{D}$, 
where $\lambda$ is the relative weight between the contributions of the edges and the nodes.

Under the chosen metric, the geodesic or the shortest path between two  points in ${\cal A}$
can be written as a set of geodesics in ${\cal S}$ between the 
corresponding components. That is, for any $A_1, A_2 \in {\cal A}$, the geodesic $\alpha: [0,1] \to {\cal A}$ 
consists of components $\alpha = \{\alpha_{ij} \}$ 
given by $\alpha_{ij}: [0,1] \to {\cal S}$, a uniform-speed geodesic path in $\mathcal{S}$ between $a_{ij}^1$ and $a_{ij}^2$. 
Note that this solution also provides an optimal registration of points across edges $a_{ij}^1$ and $a_{ij}^2$. 
Also, if needed, one can 
linearly interpolate between the attributes of the corresponding nodes, to provide node attributes for 
the intermediate graphs.

The main issue in this comparison is that 
the indexing of nodes in the graphs $A_1$ and $A_2$ is arbitrary and, thus, corresponding matching of edges
$a_{ij}^1$ with $a_{ij}^2$ is also arbitrary. 
To illustrate this point,  Fig. \ref{fig:node_order} shows the same graphical object 
several times and imposes a different node indexing every time. 
In order to provide a reasonable comparison of these graphs using the metric $d_a$, 
one has to reorder nodes every time we compare any two graphs.
Otherwise, we will end up with large $d_a$ distances between these graphs despite having the same structure. 
We shall use the permutation group to implement this reordering. This permutation is similar to the work by \cite{jain2009structure,jain2011graph} on 
graphs with Euclidean attributes.
A permutation matrix is a matrix that has exactly one 1 in each row and column, with all the other entries being zero. 
Let $\mathcal{P}$ be the group of all $n \times n$ permutation matrices with group operation being matrix multiplication and identity element being the $n \times n$ identity matrix.
We define the action of $\mathcal{P}$ on $\mathcal{B}$ as:
\begin{equation}
\mathcal{P} \times \mathcal{B} \rightarrow \mathcal{B}, P \star (A,u) = (P \cdot A \cdot P^T, Pu) \ .
\label{eqn:action}
\end{equation}
Since each entry of $A$ is an element of the shape space ${\cal S}$, this 
group action denotes a permutations of {\it shapes} in that matrix.
Here $\cdot$ implies a permutation of entries of $A$ according to the nonzero 
elements of $P$. The action $P \star A$ results in the swapping of rows and columns 
of $A$. The elements of the node attribute vector $u$ are permuted according to $P$ 
in the same way. Under the chosen metric $d_b = d_a + \lambda \Tr(D)$, this mapping is 
isometric.

\begin{lemma}\label{lem:isometry}
\begin{enumerate}
\item {\bf Permutation Group}: The action of ${\cal P}$ on the set ${\cal B}$ given in Eqn. \ref{eqn:action} is by isometries. 
That is, for any $P \in {\cal P}$, $A_1, A_2 \in {\cal A}$ and $u_1, u_2 \in \real^n$, we have 
$d_b((A_1,u_1), (A_2,u_2)) = d_b(P \star (A_1,u_1), P \star (A_2, u_2))$. 

\item {\bf Global Rotation Group}: The action of the rotation group $SO(d), d=2,3$ on ${\cal B}$, is given by 
$(A,u) \mapsto (O \odot A,u)$, which implies that each element of $A$ is rotated by the same $O$. 
In other words, $(O \odot A)_{ij} = Oa_{ij}$, for all $i, j$.
Note that the node attributes here are assumed to be invariant to graph rotations.
This rotation action on ${\cal B}$ is also by isometries. 
That is, for any $O \in SO(d)$ and $(A_1,u_1), (A_2,u_2) \in {\cal B}$, we have 
$d_b((A_1,u_1), (A_2, u_2)) = d_b(O \odot (A_1,u_1), O \odot (A_2,u_2))$. 
\end{enumerate}

\end{lemma}

Under the joint action of $\mathcal{P}$ and $SO(d)$, the orbit of an $(A,u) \in {\cal B}$ is given by:
$[(A,u)] = \{ O \odot (P\star (A,u)) |P \in \mathcal{P}, O \in SO(d) \}$.
Any two elements of an orbit denote the same graph shape, except that the nodes' ordering has changed
and the graph is rotated. Orbit membership defines an equivalence relation $\sim$ on  ${\cal A}$: $(A_1,u_1), (A_2,u_2) \in [A_1]$
implies 
$$
(A_1,u_1) \sim (A_2,u_2)  \Leftrightarrow \exists P \in \mathcal{P}, O \in SO(d): O \odot  (P \star (A_1,u_1))= (A_2,u_2) \ .
$$
The set of all equivalence classes forms the quotient space:
$\mathcal{G} \equiv\mathcal{B} \slash (\mathcal{P} \times SO(d)) = \{[(A,u)]|(A,u)\in \mathcal{B}\}$.
Henceforth, we will call ${\cal G}$ the {\it elastic graph shape space}.

\begin{lemma}
Since the finite-dimensional group  ${\cal P} \times SO(d)$ acts by isometries under the metric $d_b = d_a + \lambda d_v$, there is a well-defined induced metric on the quotient 
space ${\cal G}$, defined by
\begin{equation}
\begin{split}
d_g([(A_1,u_1)], [(A_2,u_2)]) & 
= \min_{P \in {\cal P}, O \in SO(d)} d_b((A_1,u_1), O \odot (P \star (A_2,u_2))) \ .
\label{eq:metric}
\end{split} 
\end{equation}
\end{lemma}
Since the Lie group ${\cal P} \times SO(d)$ is compact, the minimum in \eqref{eq:metric} exists. Note that the solution may not be unique, {\it i.e.}, their may be multiple ways to register two graphs optimally. 
If 
\[
(\hat{P},\hat{O}) \in \operatorname*{argmin}_{P \in {\cal P}, O\in SO(d)} d_b((A_1,u_1), O\odot (P \star (A_2,u_2))),
\]
then $(A_1,u_1)$ and $\hat{O}\odot  (\hat{P} \star (A_2,u_2))$ 
are considered to be {\it aligned and registered}. The shortest path between $[(A_1,u_1)]$ and $[(A_2,u_2)]$ under 
the metric $d_g$ is given by $[\alpha(t)]$ where
$\alpha: [0,1] \to {\cal B}$ is: (1) for edges a geodesic between $A_1$ and $\hat{O} \odot (\hat{P} \star A_2)$
and (2) for nodes it is a straight line between $u_1$ and $\hat{P} u_2$. 
This shortest path is a geodesic in the sense that each of its component is a geodesic in ${\cal S}$ 
between the registered edges. 

\subsection{Graph Matching}
\label{sec:graphmatching}
The problem of optimization over $\mathcal{P}$, stated in Eqn. \ref{eq:metric}, 
is an instance of a {\it generalized graph matching problem} and is the most important challenge in the proposed framework. Given an optimal $P \in {\cal P}$, 
the optimization over $SO(d)$ is straightforward, and the result is obtained using the Procrustes method. We will not discuss it further and will focus only on graph matching. 
This matching problem is, in fact, NP complete~[\cite{yan2016short}], and its global solution cannot be found in a reasonable time as the graph size increases. Instead, one uses relaxation techniques to find approximate solutions in one of several ways. 

If the adjacency matrix
$A$ were real-valued and edge similarities were measured by 
the Euclidean norm,  then the matching problem can be written as 
$\hat{P} = \argmin_{P\in\mathcal{P}}\|A_1-PA_2P^T\|=\argmax_{P\in\mathcal{P}}\Tr(A_1PA_2P^T)$. 
This particular formulation is known as 
{\it Koopmans-Beckmann's} quadratic assignment programming (QAP) problem~[\cite{koopmans1957assignment}].
In this case, one can use an existing relaxation
solution for approximating the optimal registration~[\cite{caelli2004eigenspace,liu2012extended,umeyama1988eigendecomposition,vogelstein2015fast}].

However, when elements of $\mathcal{A}$ belong to a more general 
metric space, {\it e.g.}, shape space $\mathcal{S}$, 
some of the previous solutions are not applicable.
Instead, the problem can be rephrased as 
$\hat{P} = \argmax_{P \in \mathcal{P}} \vect{(P)}^TK\vect{(P)}$, 
where $ \vect{(P)}$ denotes a concatenation of elements of $P$ 
in a column vector. 
Here $K \in \real^{n^2 \times n^2}$, called an {\it affinity matrix}, has the following structure. 
Suppose $A_1$ has node index $a, b$, etc. and $A_2$ has node index  $i, j$, etc. Then, 
\begin{itemize}
\item 
the diagonal entries $k_{aiai}$ measure the affinity between node $a$ of $A_1$ and 
node $i$ of $A_2$, and

\item the off-diagonal entries $k_{aibj}$ measures the affinity between edge $ab$ of $A_1$ and edge $ij$ of $A_2$.
\end{itemize}
In this paper we use the shape similarity between two edges, while modding out the re-parametrization group, as affinity: 
$$
k_{aibj}=\begin{cases}
0,\  \text{if either $a^1_{ab}$ or $a^2_{ij}$ is null} \\
\sup_{\gamma}\langle q_1, O(q_2 \circ \gamma)\sqrt{\dot{\gamma}} \rangle,\ \text{otherwise}
\end{cases} \ .
$$
Note that this $\sup$ over $\gamma$ is related to the infimum over $\gamma$ in the definition of $d_s$ because: 
$$
\| q_1 - O (q_2*\gamma)\|^2 = \|q_1\|^2 + \|q_2\|^2 - \langle q_1, O(q_2 \circ \gamma)\sqrt{\dot{\gamma}} \rangle\ .
$$
Here $q_1, q_2$ denote the SRVFs of the edges $ab$ of $A_1$ and $ij$ of $A_2$ and
$\gamma$ is a diffeomorphic reparameterization and the
supremum\textcolor{green}{,} computed using the Dynamic Programming Algorithm (DPA)~[\cite{srivastava2010shape}].
The matrix $O \in SO(d)$ is the global rotation matrix that is 
used to rotationally align two elastic graphs. (The same $O$ matrix is applied to all the edges of the second graph and 
minimized using the Procrustes rotation.)
When $\lambda > 0$ and node attributes are involved in shape analysis, we can set 
$k_{aiai} = \lambda \exp(-\frac{|u_1(a)-u_2(i)|^2}{2\sigma^2})$. In practice, the choice of 
$\lambda$ and $\sigma$ are an issue and one can use learning or
cross-validation to estimate suitable values.

The resulting formulation is called the {\it Lawler's QAP problem}~[\cite{lawler1963quadratic}].
Koopmans-Beckmann's QAP is a particular case of Lawler's QAP.
To solve for Lawler's QAP,  at least approximately, there are several algorithms available
~[\cite{cour2007balanced,gold1996graduated,leordeanu2005spectral,
leordeanu2009integer,zanfir2018deep,zhou2012factorized,zhou2015factorized}].
In this paper, we use the well-known {\it factorized graph matching} (FGM) algorithm~[\cite{zhou2015factorized}] to match elastic graphs.
As mentioned previously, we need to optimize edge affinity over
both orientation-preserving and reversing diffeomorphisms, to allow for the possibility of an orientation-reserving registration (the original DPA is only applicable for orientation-preserving diffeomorphisms). 
We can work around it using the directed FGM. We treat each edge $a_{ij}$ in the graph as two directed edges: one from $i$ to $j$ and one from $j$ to $i$. The edge affinity between any two edges is evaluated four times (two directions of one edge times two directions of another edge) using the original DPA. In practice, we calculate the affinities only twice and use each value twice.

\subsection{Introducing Null Nodes}
Thus far, we have assumed that the graphs being matched are all of the same size (in terms of the number of nodes). 
For graphs $G_1$ and $G_2$, with different numbers of nodes $n_1$ and $n_2$, we 
can append them with $n_2$ and $n_1$ null nodes, respectively, to bring them each to the same size $n_1 + n_2$. 
This addition of null nodes can be done even when they are the same size, as it can help improve graph matching. 
This way, the original (real) nodes of both $G_1$ and $G_2$ can potentially be registered to null nodes in the other graph and bring down the matching cost.

When we add null nodes to a graph, we need to assign attributes to these extra nodes and edges thus created. We set the edges connected to null nodes to have a value
$\mathbf{0} \in {\cal S}$. In other words, 
we extend the adjacency matrices of the two graphs from $A_i$ to $\tilde{A}_i$ as shown below. 
For the null nodes, we do not assign the node attributes explicitly. Instead, we extend the node distance matrix $D$ to $\tilde{D}$ by setting the entries corresponding to the null nodes to be zero. This setting implies that the null nodes' attributes are equal to those of the nodes they are being matched to in the other graph. This choice ensures that the attributes of null nodes
do not contribute to the matching cost. The extended matrices are constructed as follows:
\\ 

\raisebox{0.2in}{$\tilde{A}_1 =$}~~~ \begin{tikzpicture}[decoration=brace]
    \matrix (m) [matrix of math nodes,left delimiter=[,right delimiter={]}] {
        A_1 & ~ &{\bf 0} \\
        ~ & ~ &~\\
        {\bf 0} &  ~ &{\bf 0} \\
          };
    \draw[decorate,transform canvas={xshift=-1.0em},thick] (m-1-1.south west) -- node[left=2pt] {$n_1$} (m-1-1.north west);
     \draw[decorate,transform canvas={xshift=-1.3em},thick] (m-3-1.south west) -- node[left=2pt] {$n_2$} (m-3-1.north west);
     \draw[decorate,transform canvas={yshift=0.5em},thick] (m-1-1.north west) -- node[above=2pt] {$n_1$} (m-1-1.north east);
    \draw[decorate,transform canvas={yshift=0.5em},thick] (m-1-3.north west) -- node[above=2pt] {$n_2$} (m-1-3.north east);
\end{tikzpicture}, 
\raisebox{0.2in}{$\tilde{A}_2 =$}~~~ \begin{tikzpicture}[decoration=brace]
    \matrix (m) [matrix of math nodes,left delimiter=[,right delimiter={]}] {
        A_2 & ~ &{\bf 0} \\
        ~ & ~ &~\\
        {\bf 0} &  ~ &{\bf 0} \\
          };
    \draw[decorate,transform canvas={xshift=-1.0em},thick] (m-1-1.south west) -- node[left=2pt] {$n_2$} (m-1-1.north west);
     \draw[decorate,transform canvas={xshift=-1.3em},thick] (m-3-1.south west) -- node[left=2pt] {$n_1$} (m-3-1.north west);
     \draw[decorate,transform canvas={yshift=0.5em},thick] (m-1-1.north west) -- node[above=2pt] {$n_2$} (m-1-1.north east);
    \draw[decorate,transform canvas={yshift=0.5em},thick] (m-1-3.north west) -- node[above=2pt] {$n_1$} (m-1-3.north east);
\end{tikzpicture},
\raisebox{0.2in}{$\tilde{D} =$}~~~ \begin{tikzpicture}[decoration=brace]
    \matrix (m) [matrix of math nodes,left delimiter=[,right delimiter={]}] {
        D & ~ &{\bf 0} \\
        ~ & ~ &~\\
        {\bf 0} &  ~ &{\bf 0} \\
          };
    \draw[decorate,transform canvas={xshift=-1.0em},thick] (m-1-1.south west) -- node[left=2pt] {$n_1$} (m-1-1.north west);
     \draw[decorate,transform canvas={xshift=-1.3em},thick] (m-3-1.south west) -- node[left=2pt] {$n_2$} (m-3-1.north west);
     \draw[decorate,transform canvas={yshift=0.5em},thick] (m-1-1.north west) -- node[above=2pt] {$n_2$} (m-1-1.north east);
    \draw[decorate,transform canvas={yshift=0.5em},thick] (m-1-3.north west) -- node[above=2pt] {$n_1$} (m-1-3.north east);
\end{tikzpicture}.
\\
Once we have extended the graphs using the null nodes, we apply the same matching 
procedure as before. In terms of displaying the extended graphs, we do not display null nodes
matched with null nodes. The null nodes matched with real nodes are assigned the same
coordinates, as their matched counterparts, for display purposes.

\begin{figure}
\begin{center}
\begin{tabular}{|@{}c@{} @{}c@{} |  @{}c@{}  @{}c@{}|}
\hline
\includegraphics[height=1in]{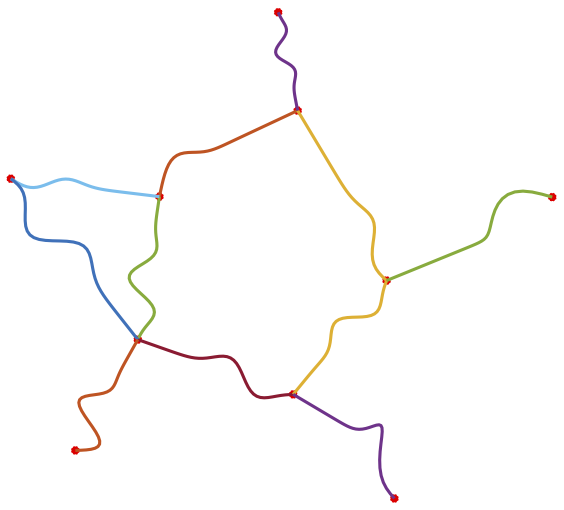} &
\includegraphics[height=1in]{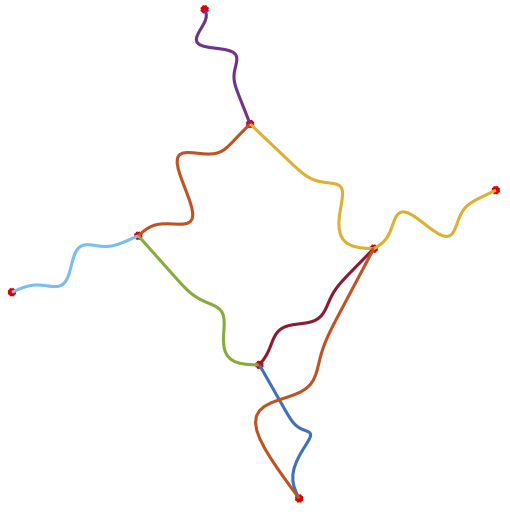}  &
\includegraphics[height=1in]{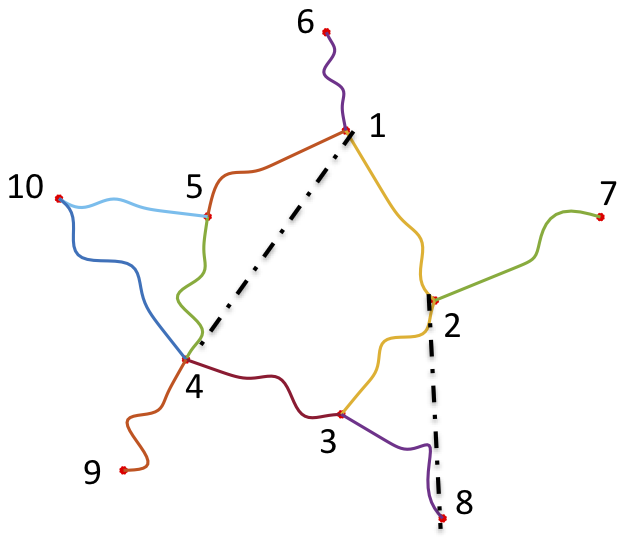} &
\includegraphics[height=1in]{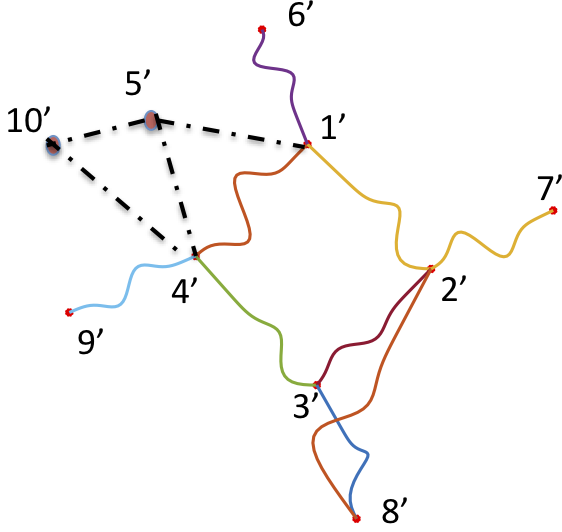} \\
Graph $G_1$ & Graph $G_2$ & \multicolumn{2}{c|}{Matched graphs padded with null nodes/edges} \\
\hline
\end{tabular}
\caption{Labeling and addition of null nodes in order to facilitate registration between 
two graphs.} \label{fig:null-nodes}
\end{center}
\end{figure}

Figure~\ref{fig:null-nodes} shows an example of this idea. The left side shows two graphs that have
different number of nodes and edges, but still seem to have a common structure. On the right side we label the
nodes to show a particular matching of these graphs. The matched nodes are labelled $1-1'$, 
$2-2'$, and so on. We obtain this matching by adding two null 
nodes: $5'$ and $10'$, and corresponding null edges $1' \leftrightarrow 5'$, $10' \leftrightarrow 5'$, 
$10' \leftrightarrow 4'$, $5' \leftrightarrow 4'$
in $G_2$, and the null edges $1 \leftrightarrow 4$, 
$2 \leftrightarrow 8$ in $G_1$. We emphasize that the matching between graphs is performed at the node level and is one-to-one. This framework does not allow nodes to split or merge when registering across graphs (as in, \textit{e.g.}, \cite{Chowdhury_2020_CVPR_Workshops}). Consequently, the edges (where real or null) are also matched in a one-to-one fashion. Once again, there is no split or merging of edges allowed when matching graphs.
For instance, in Fig.~\ref{fig:null-nodes}, the edge $1 \leftrightarrow 4$ can only be matched with a single edge, say 
$1'  \leftrightarrow 4'$, but can not be matched with the pair say $1'  \leftrightarrow 5'  \leftrightarrow 4'$.

\begin{figure}
\begin{center}
\begin{tabular}{@{}c@{}|@{}c@{}}
\hline
Geodesic in ${\cal B}$ & Geodesic in ${\cal G}$ \\
\hline
\includegraphics[height=0.45in]{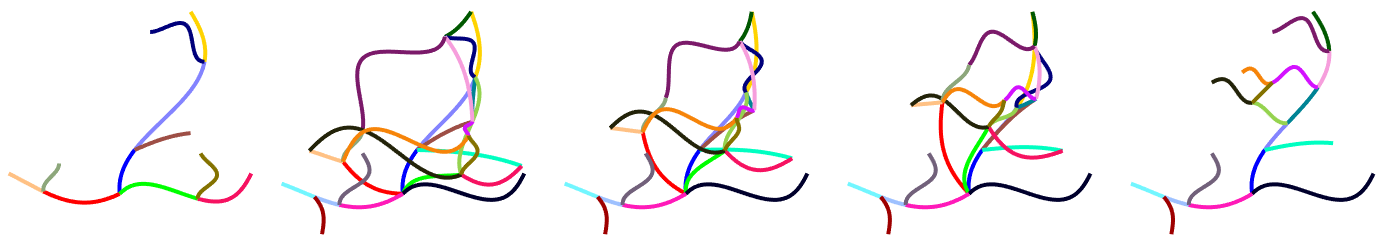} &
\includegraphics[height=0.45in]{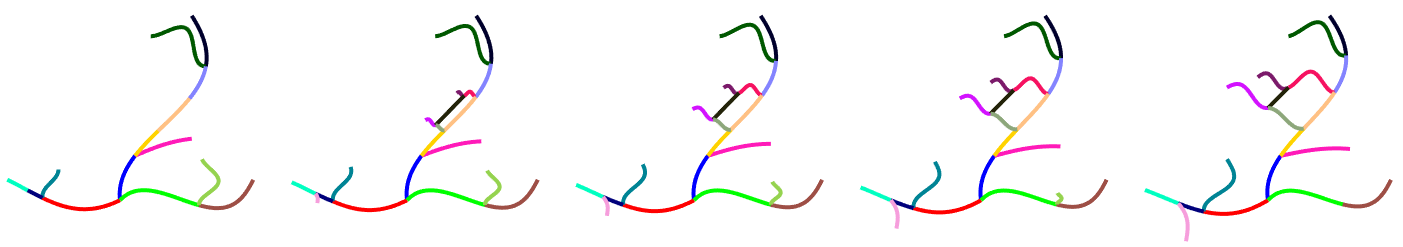}\\
\hline
\includegraphics[height=0.55in]{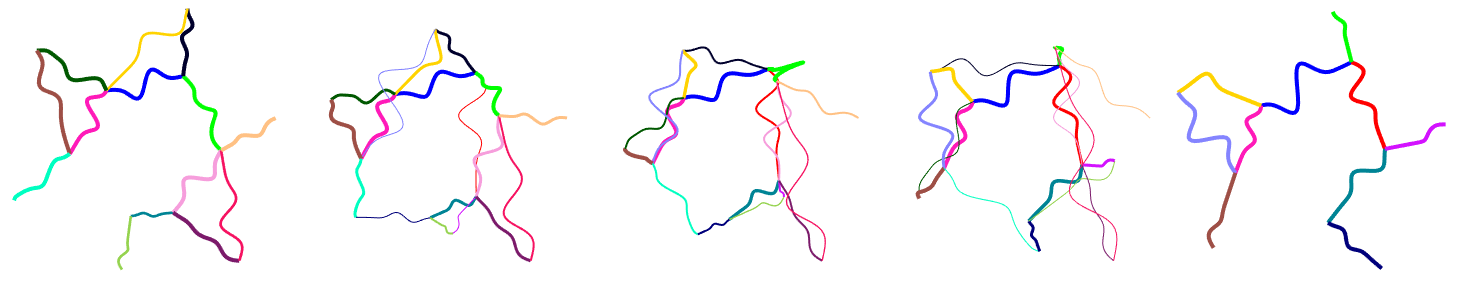} &
\includegraphics[height=0.55in]{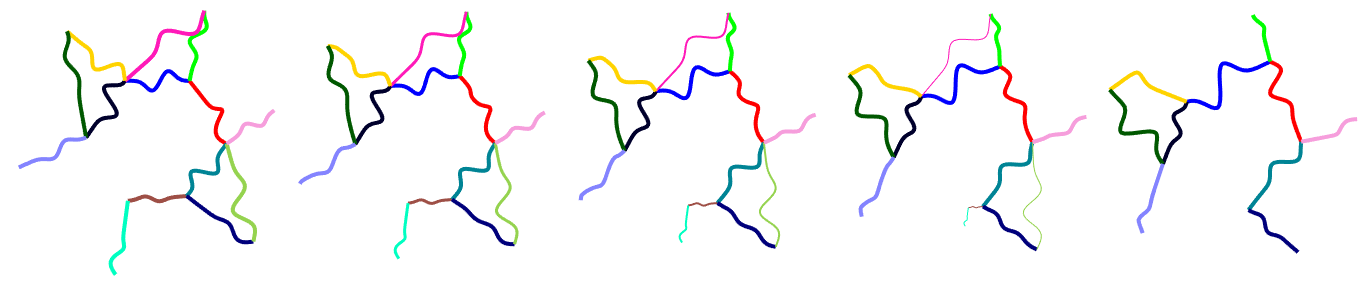}\\
\hline
\includegraphics[height=0.5in]{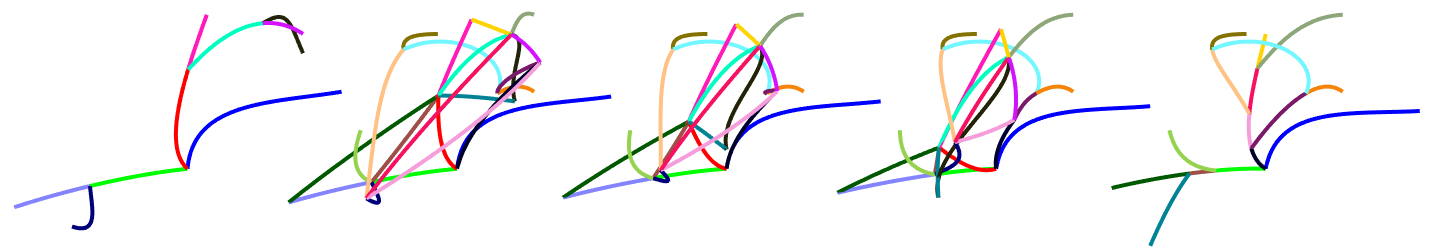} &
\includegraphics[height=0.5in]{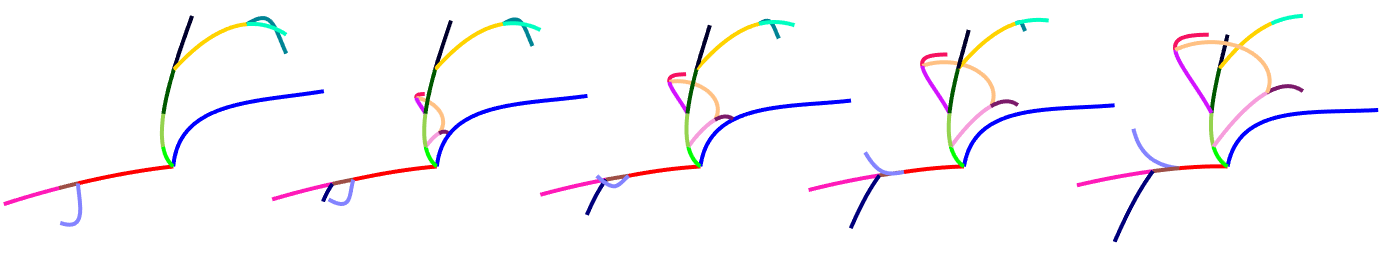}\\
\hline
\includegraphics[height=0.75in]{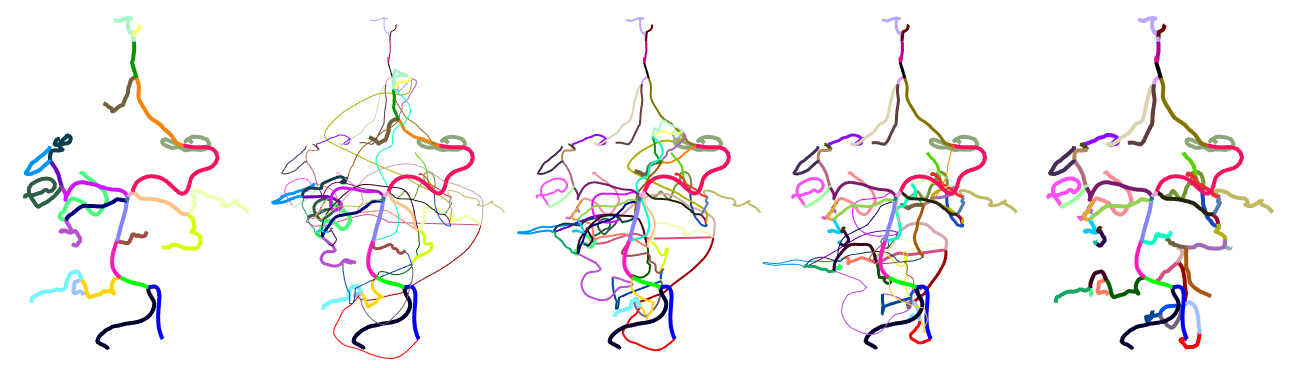}&
\includegraphics[height=0.75in]{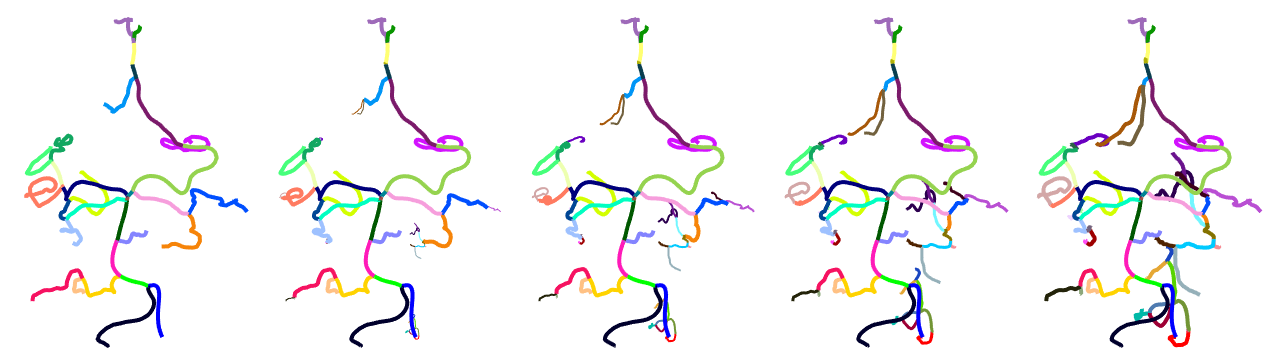}\\
\hline
\end{tabular}
\end{center}
\caption{Geodesic between graphs in space ${\cal B}$ (left) and the ${\cal G}$ (right).
Coloring is used to depict registration of edges across graphs. The top two rows show 2D graphs
while bottom two use 3D graphs. (GIF animations of these geodesics are provided with
the supplementary material~[\cite{supplement}].)}
\label{fig:geo_simu}
\end{figure}

Once we match the two graphs, we can compute geodesics between them by interpolating between the corresponding
nodes and edges (according to their respective metrics). 
We present four illustrative examples in 
Fig. \ref{fig:geo_simu}. 
The two graphs $G_1$ and $G_2$ drawn as the first and the last graphs in each picture in each sequence. 
Additionally, we show a sequence of shapes along the geodesic path between them in two different spaces -- ${\cal B}$ and ${\cal G}$, 
{\it i.e.}, with
arbitrary registration and with optimal registration. The deformations between registered graphs, associated with geodesics in ${\cal G}$,  look much more natural than those in ${\cal B}$.  The edge features are
preserved better in the intermediate graphs along the geodesics in ${\cal G}$.  The edges
in one graph that are unmatched in the other graph either disappear or appear along the geodesic. 
We can display this effect using either thickening/thinning or shrinking/growing of edges. 
For internal edges, the change in thickness seems more appropriate, but for the terminal edges, {\it i.e.,} edges that have a node with only one edge attached to it, 
it seems more natural to shorten (lengthwise) a non-null edge into a null edge. 
Note that the framework currently does not utilize any information about edge thickness in shape comparisons. This
property is manipulated only for display purposes.   
The top two rows are for 2D graphs, while the remaining examples are 3D graphs. 
As mentioned earlier,  the points along registered edges of graphs are registered while computing $d_s$.
Thus, we reach a dense (complete) registration of parts across the two graphs, making the deformations
appear more natural.

\section{Shape Summaries of Elastic Graphs}
\label{sec:shapesummary}

We are interested in tools that facilitate statistical inferences for the given BAN dataset. These inferences include classification, clustering, hypothesis testing, and modeling. The use of a metric structure to compute summaries of shapes of graphs is of great importance in these analyses. We will use the earlier metric structure to define and compute shape statistics -- such as mean, covariance, and PCA -- of given graph data. Further, we will use these representations to perform dimension reduction and hypothesis testing. 

\subsection{Mean Graph Shapes}
Given a set of graph shapes $\{ [A_i] \in {\cal G}, i = 1,2,\dots,m\}$, we define their mean graph shape 
to be: 
$$
[A_{\mu}] = \argmin_{[A] \in {\cal G}} \left( \sum_{i=1}^m d_g([A],[A_i])^2 \right)\ ,
$$
where $d_g$ is as defined in Eqn.~\ref{eq:metric}. This construction has been named
Fr\'{e}chet, Karcher, or  intrinsic mean interchangeably. 
There are at least two different ways of computing this mean. One relies on the gradient of the cost function 
in this optimization, and the other relies on finding a sequence of geodesic paths. In the following, we focus on computing 
the means of the edge attributes, since the computation of node averages is relatively straightforward.
\\

\noindent {\bf Method 1-- Gradient Approach}:  
Algorithm \ref{algo:mean} outlines a gradient-based approach for computing the mean shape. 
This gradient solution is a local minimum of the cost functional and does not guarantee a global minimizer.

\begin{algorithm}
\caption{Graph Mean in ${\cal G}$}
\label{algo:mean}
\begin{flushleft}
Given adjacency matrices $A_{i}$, $i=1,..,m$:
\end{flushleft}
\begin{algorithmic}[1]
\State Initialize a mean template $A_{\mu}$ (e.g., the largest graph).
\item Rotationally align $A_i$s to $A_{\mu}$ using the Procrustes method.
\State Match $A_{i}$ to $A_{\mu}$ under the permutation group ${\cal P}$ using FGM [\cite{zhou2015factorized}]
and SRVF~[\cite{srivastava2016functional}], store the matched graph shape as $A_{i}^*$, for $i = 1,.., m$.
\State Update $A_{\mu} = \frac{1}{m}\sum_{i=1}^{m} A_{i}^*$. Since the graphs are 
all registered to $A_{\mu}$, we can take 
a Euclidean average of the elements of $A_i^*$s here.
\State Repeat 2 and 3 until $\sum_{i=1}^{m} d_a(A_{i}^*,A_{\mu})^2$ converges.
\end{algorithmic}
\end{algorithm}

\begin{figure}[t]
\begin{center}
\begin{tabular}{|@{}c@{} @{}c@{} @{}c@{} @{}c@{}  @{}c@{} @{}c@{} |@{}c@{}|}
\hline
\multicolumn{6}{|c|}{Sample Shapes} & Mean Shape \\
\hline
\includegraphics[height=0.7in]{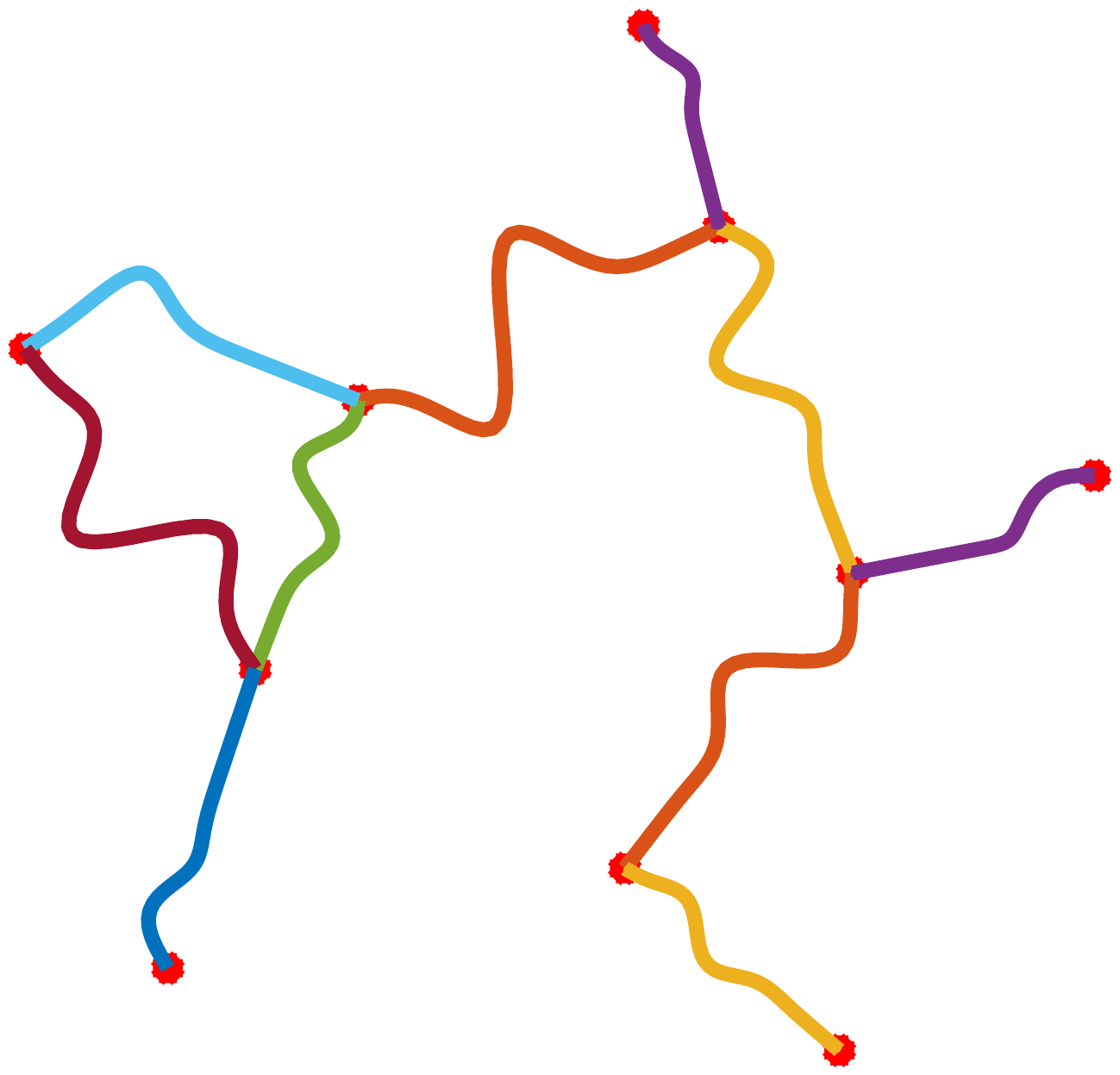} &
\includegraphics[height=0.7in]{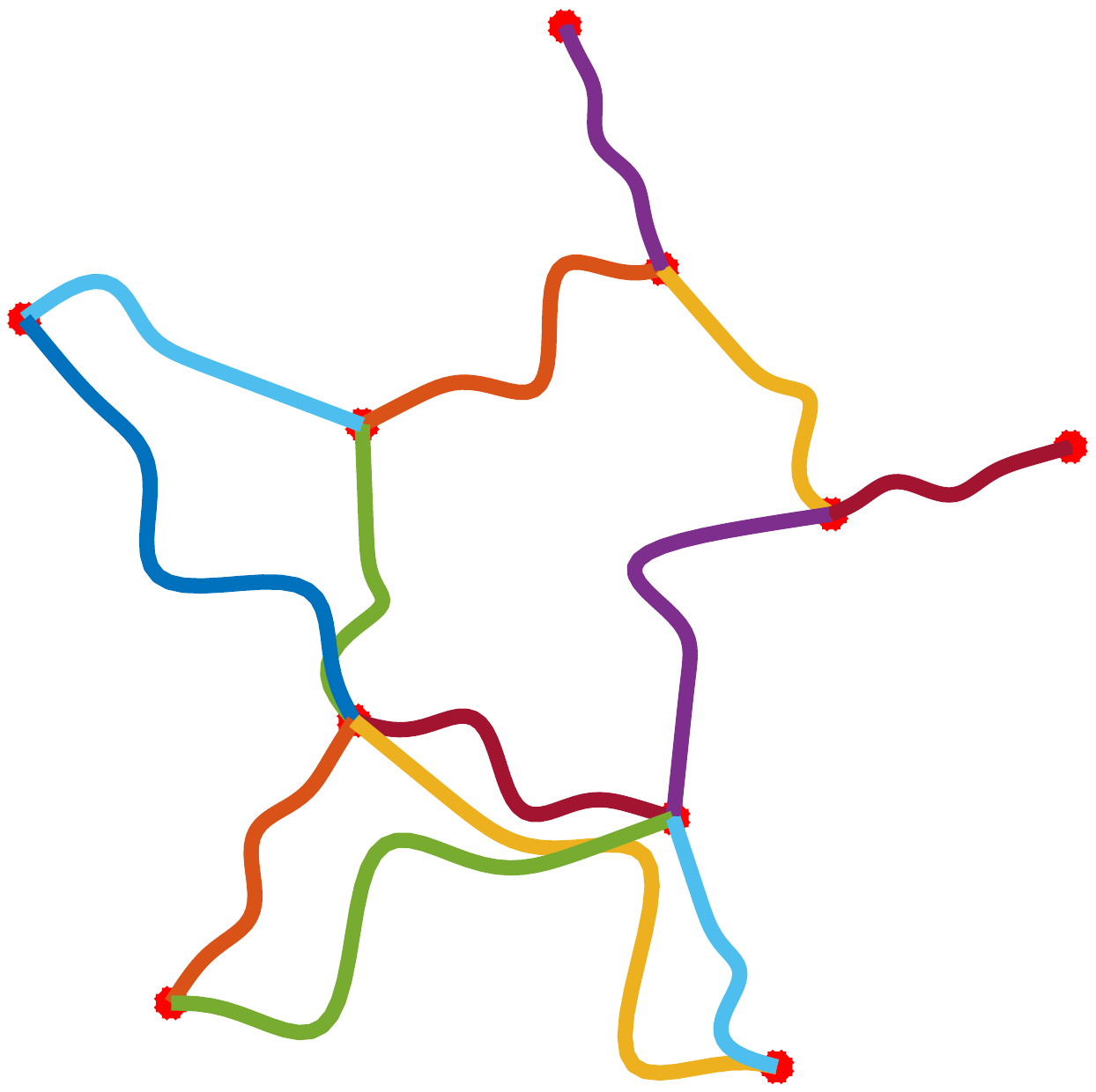} &
\includegraphics[height=0.7in]{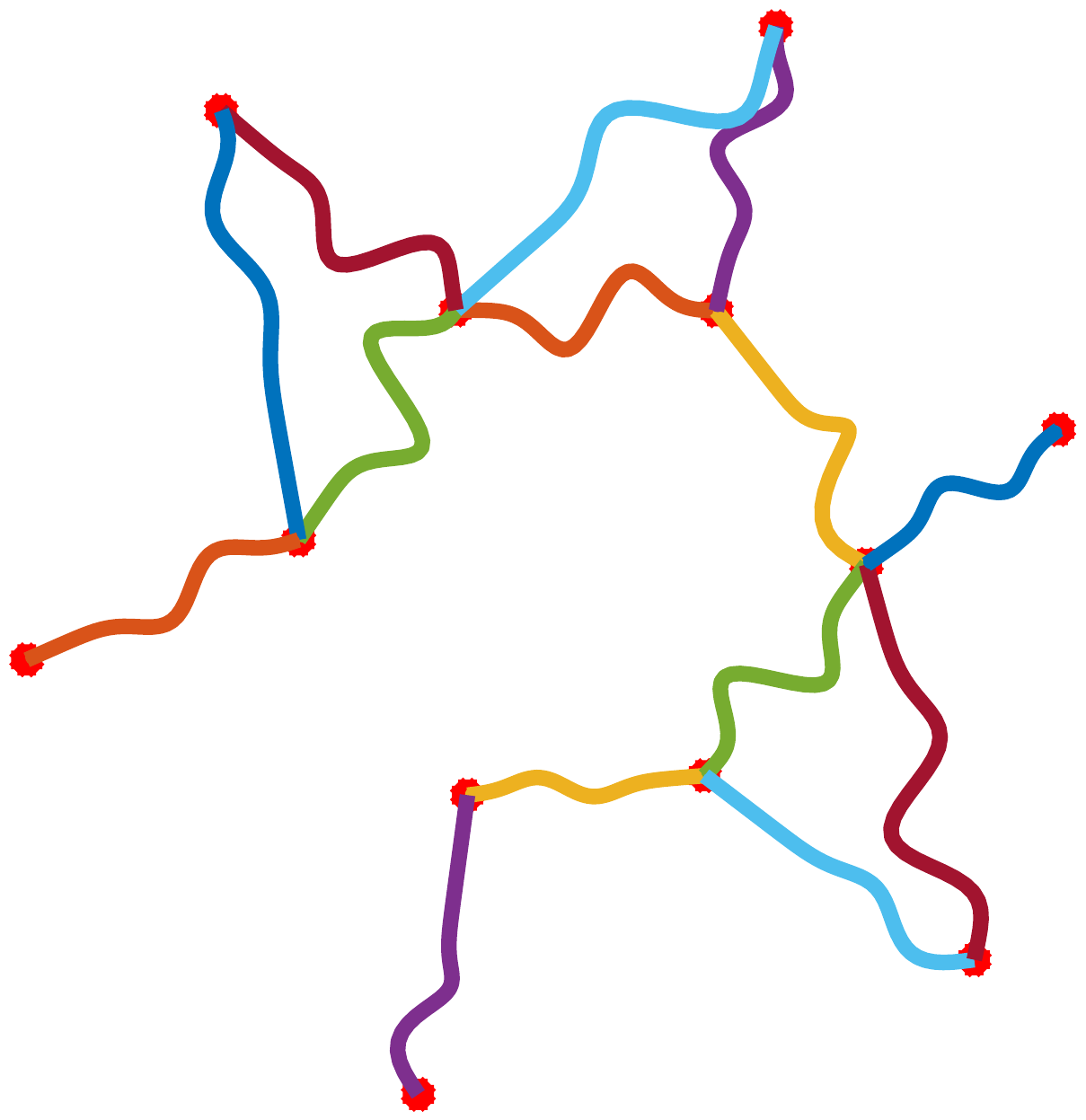} &
\includegraphics[height=0.7in]{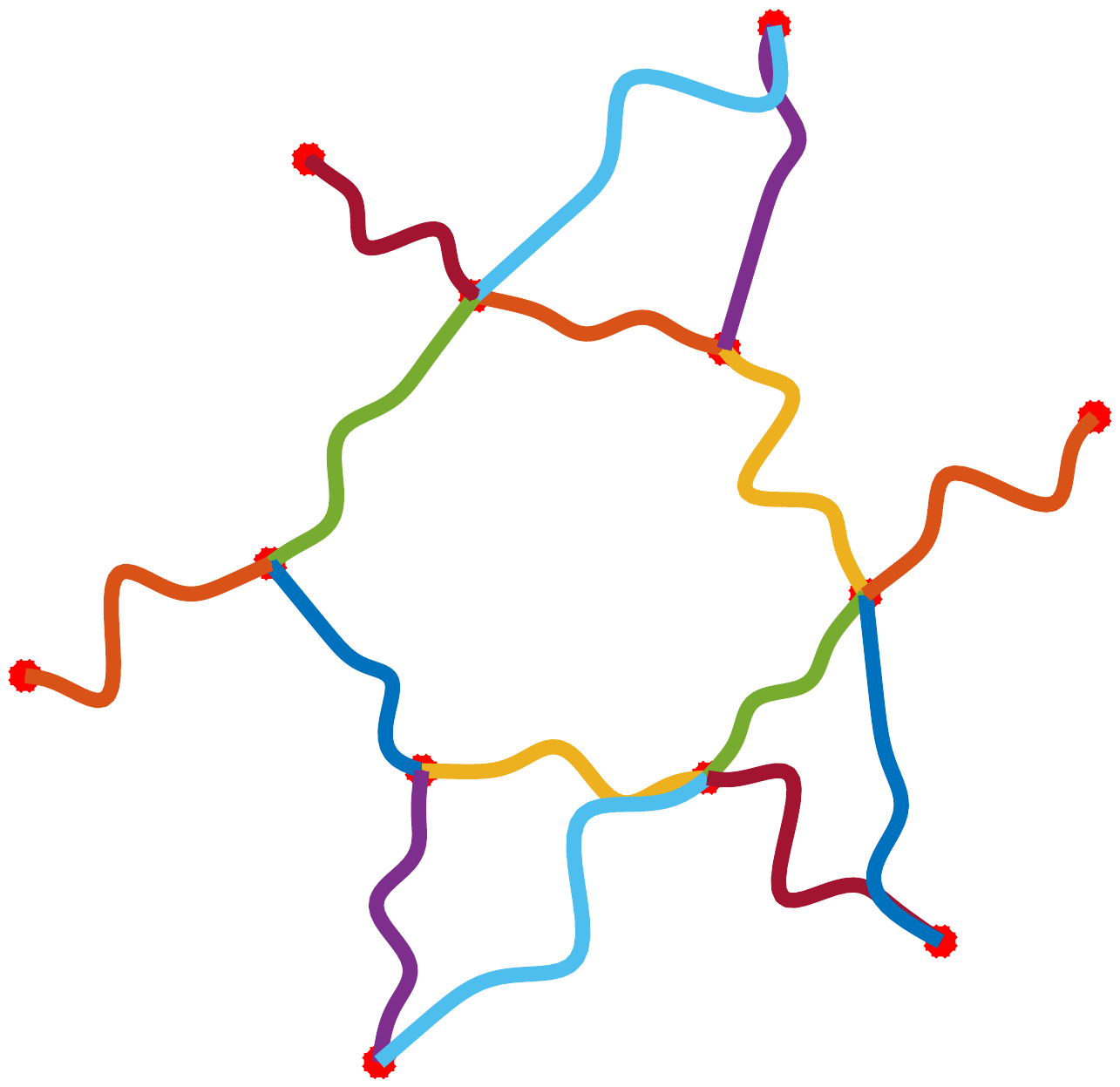} &
\includegraphics[height=0.7in]{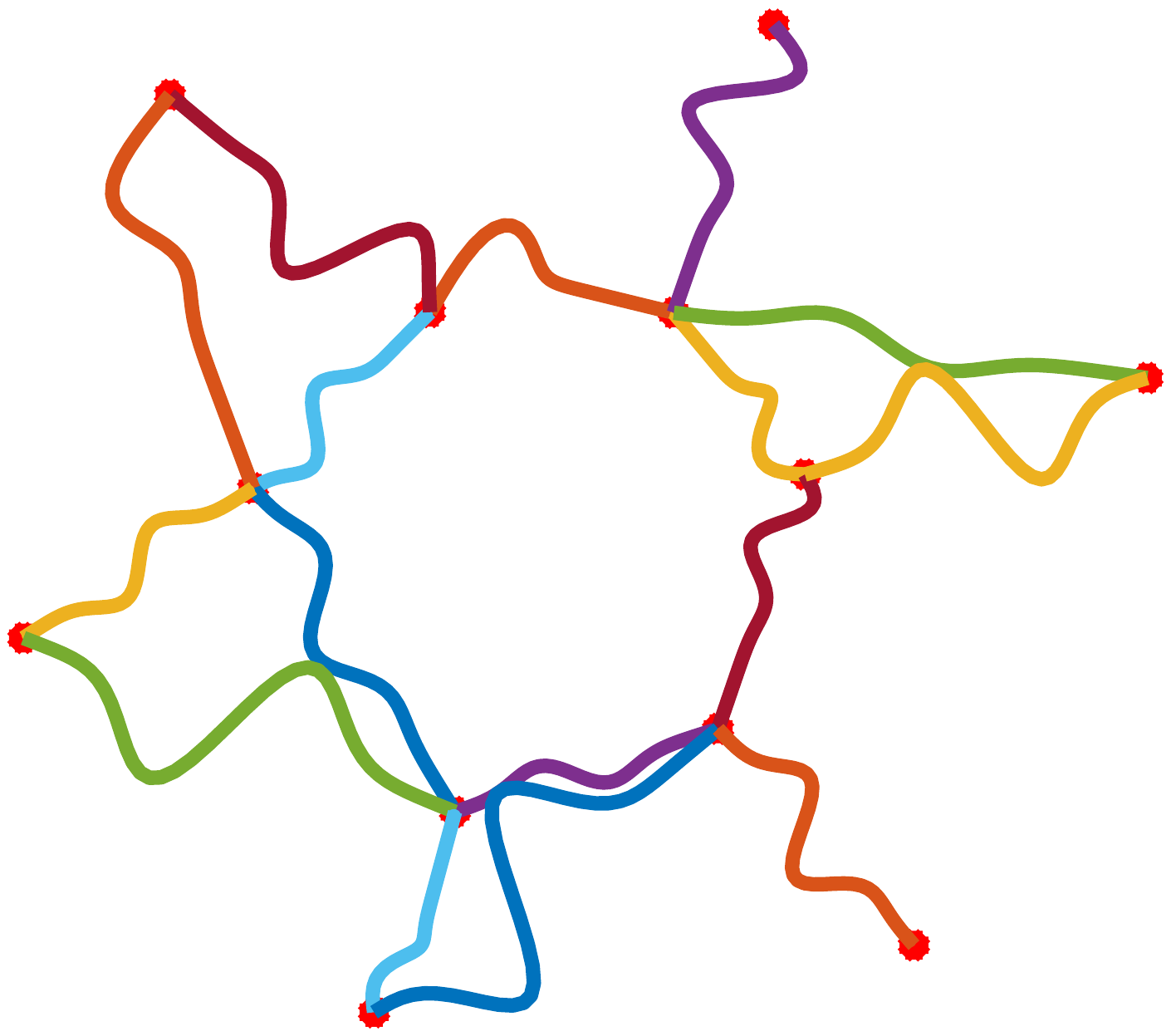} &
\includegraphics[height=0.7in]{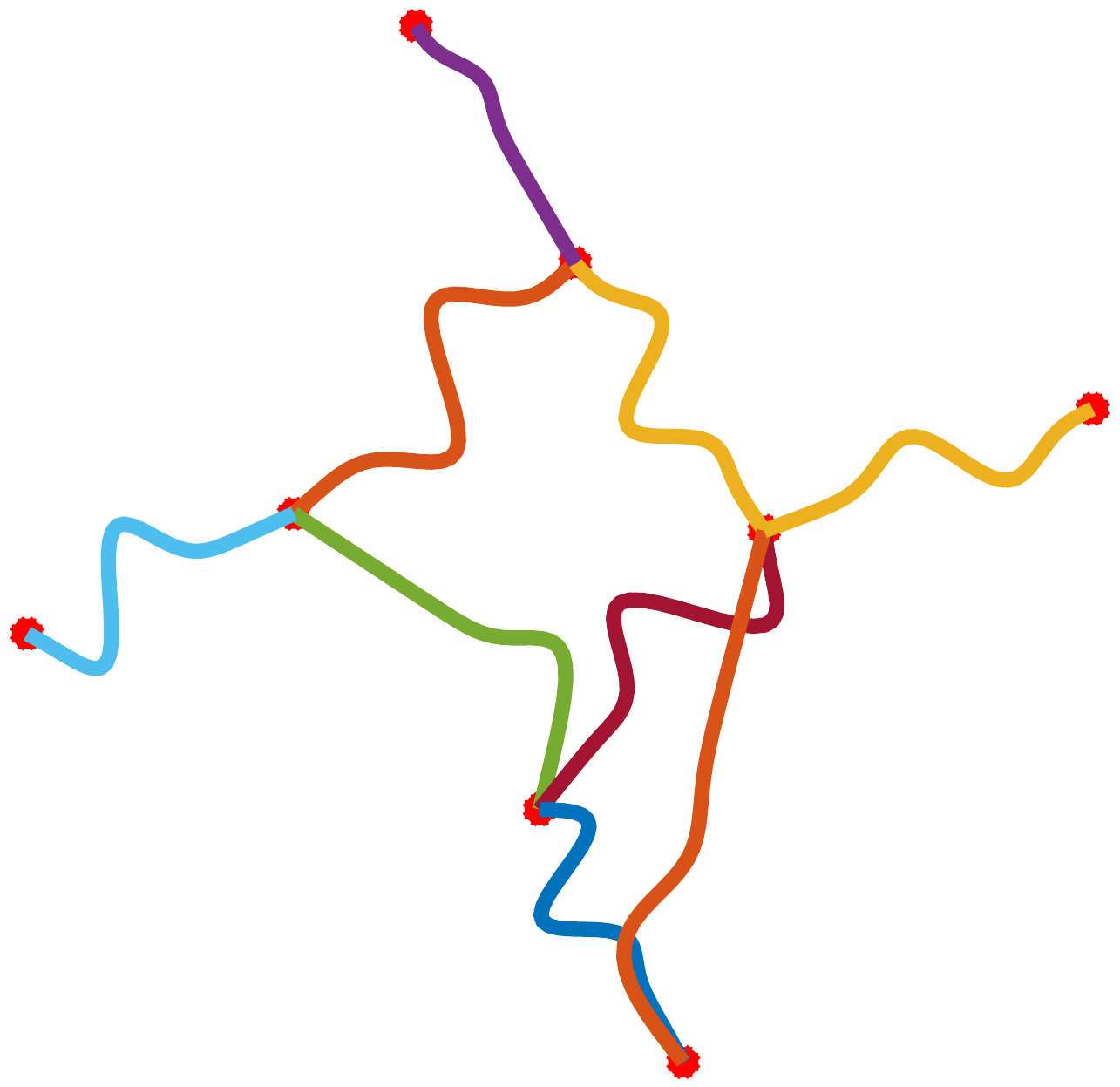} &
\includegraphics[height=0.7in]{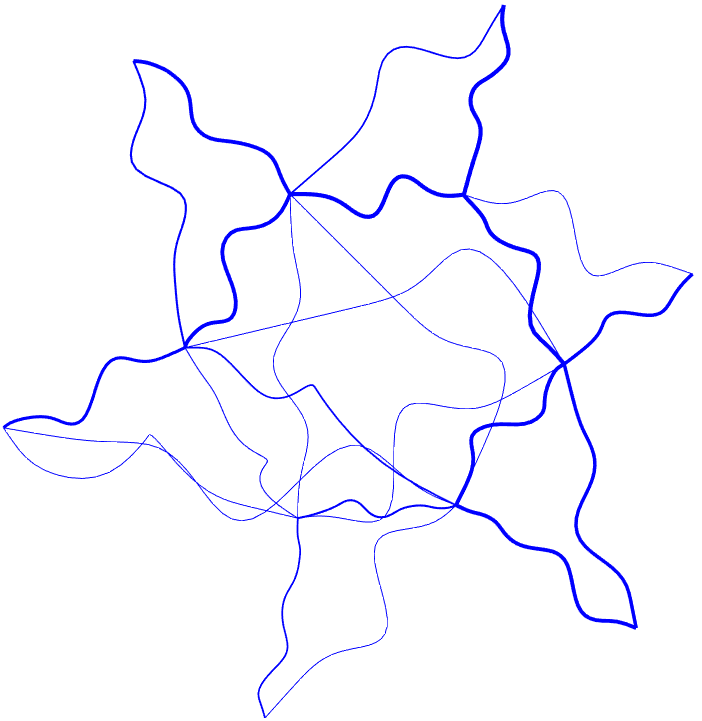} \\
 & & & & & & Method 1 \\
\includegraphics[height=0.7in]{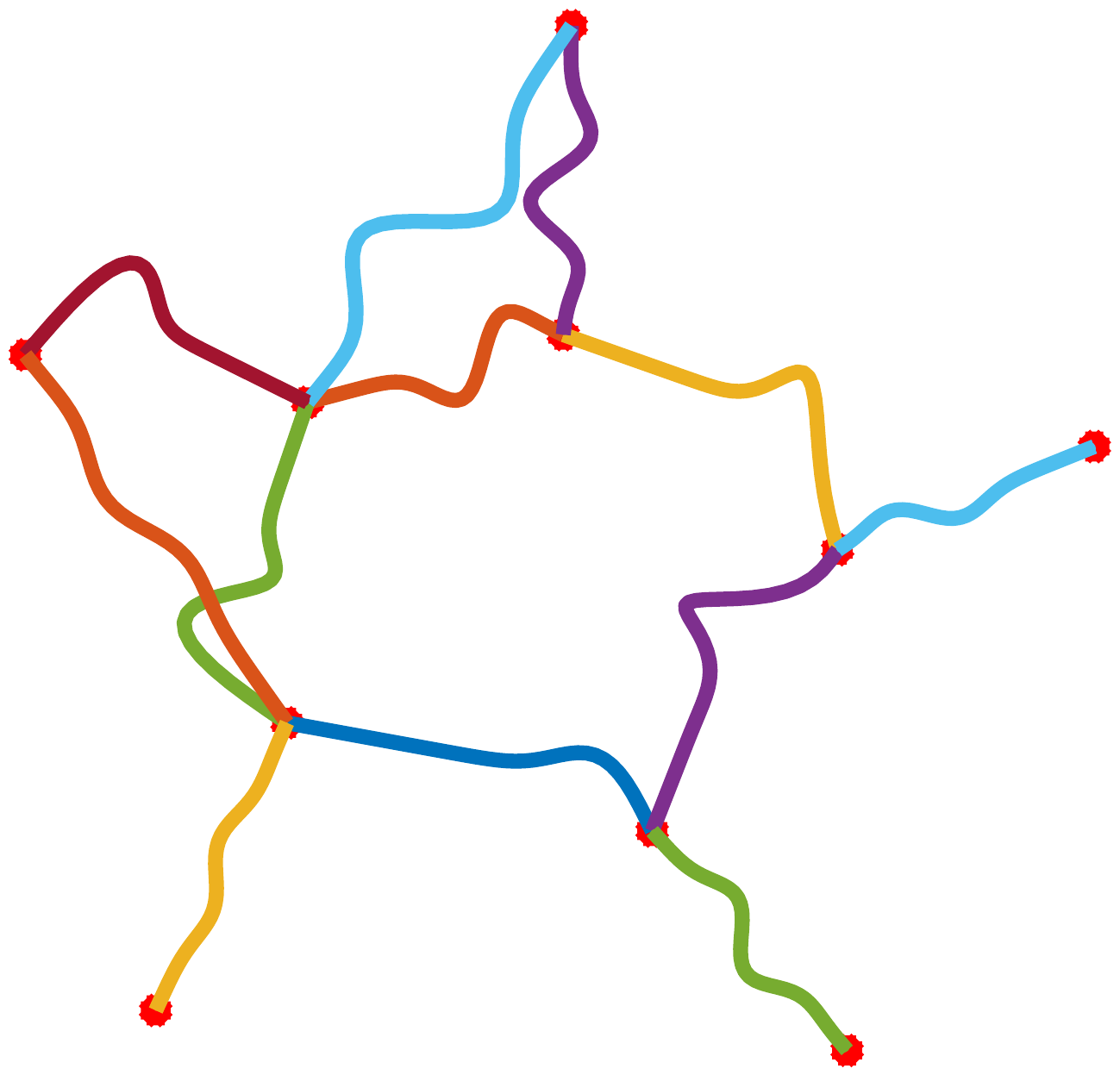} &
\includegraphics[height=0.7in]{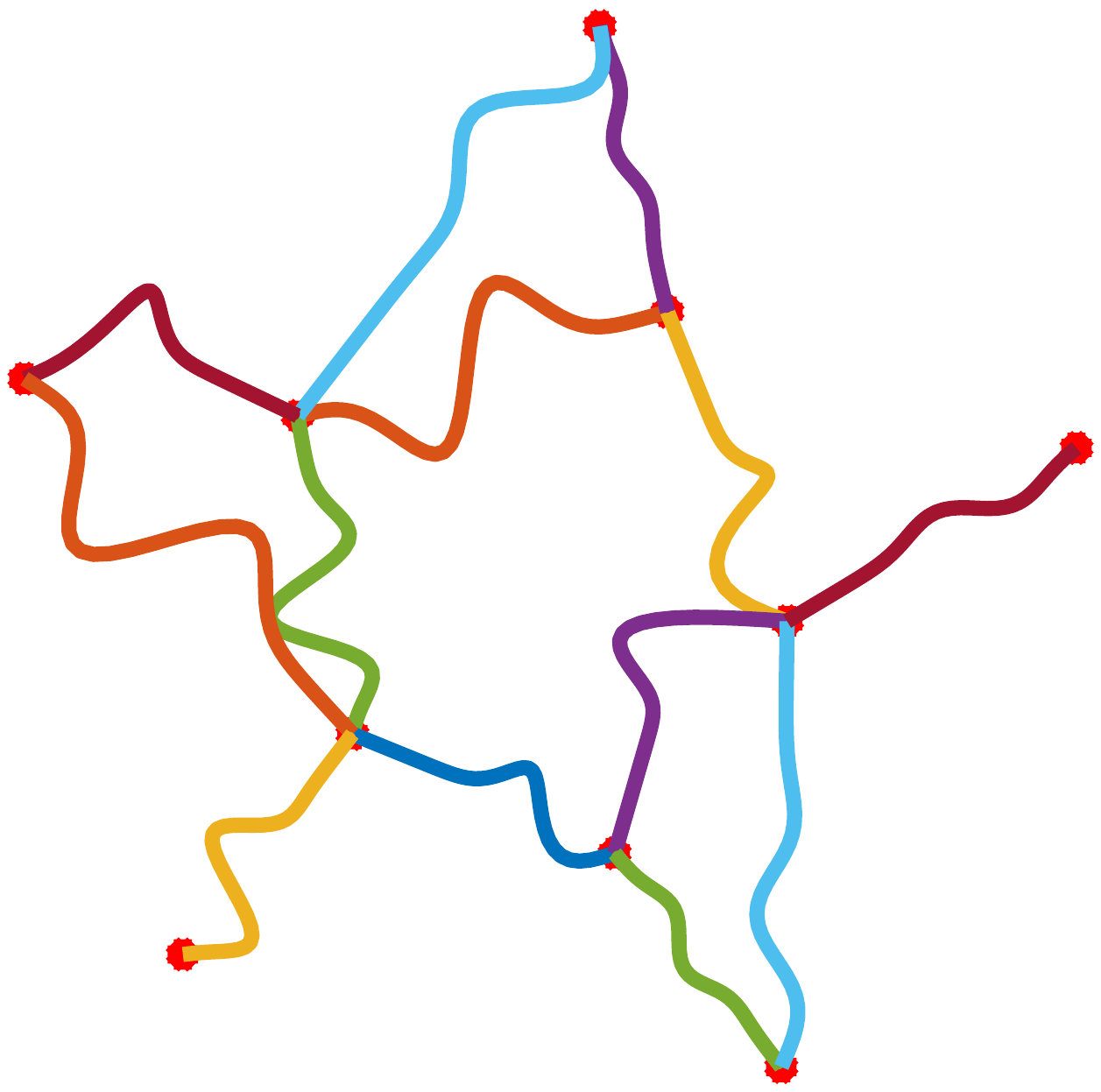}&
\includegraphics[height=0.7in]{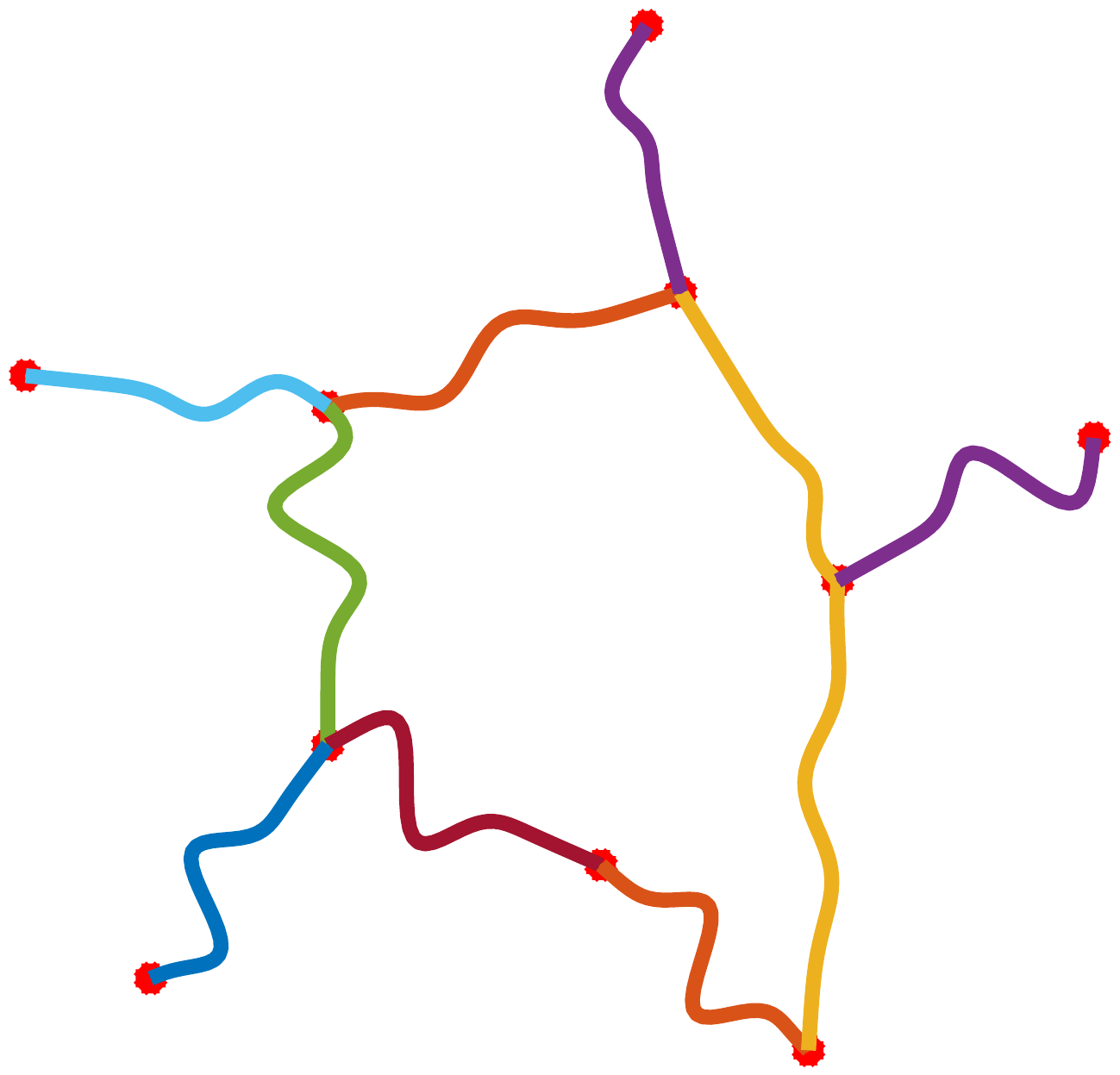} &
\includegraphics[height=0.7in]{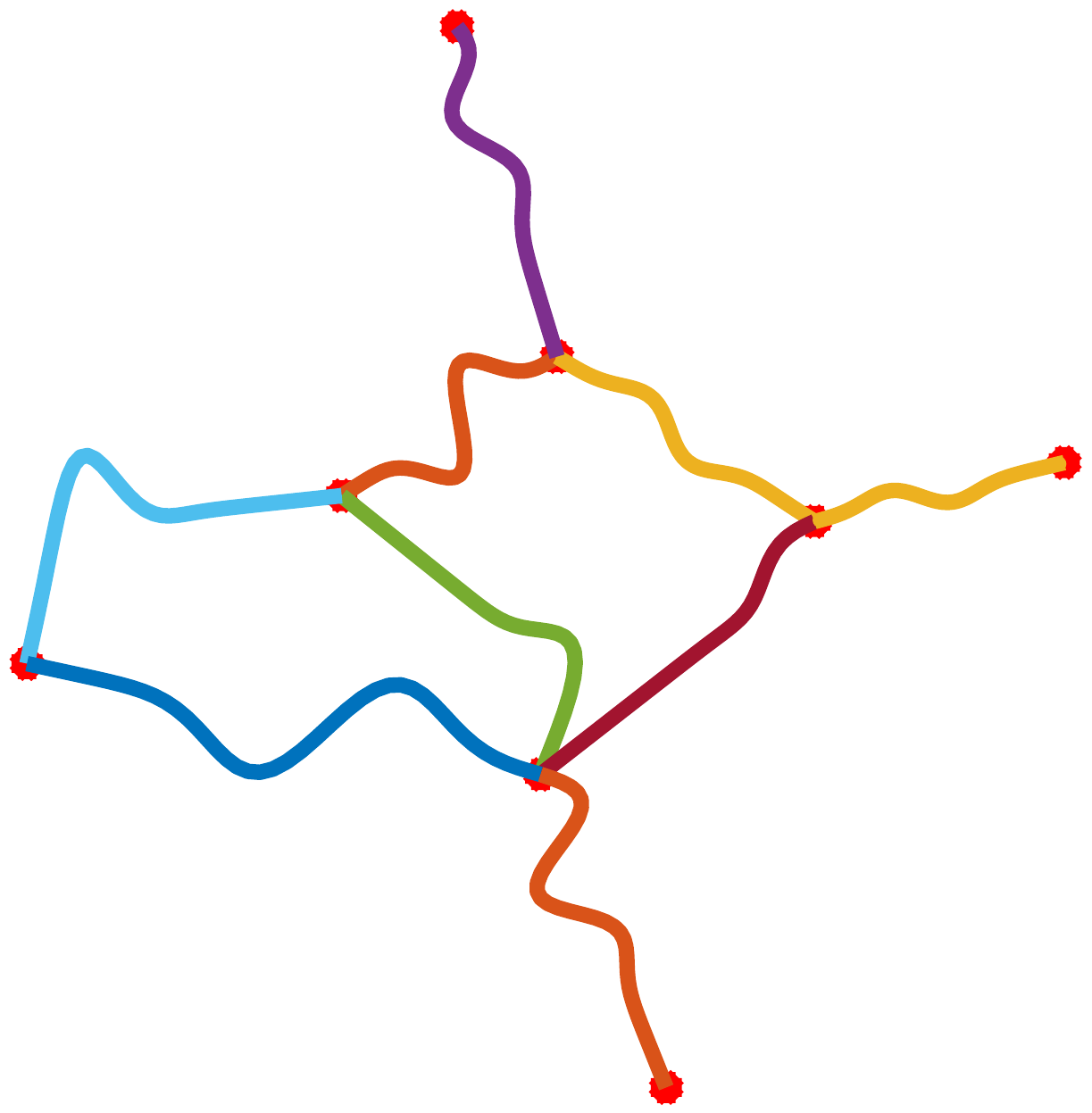} &
\includegraphics[height=0.7in]{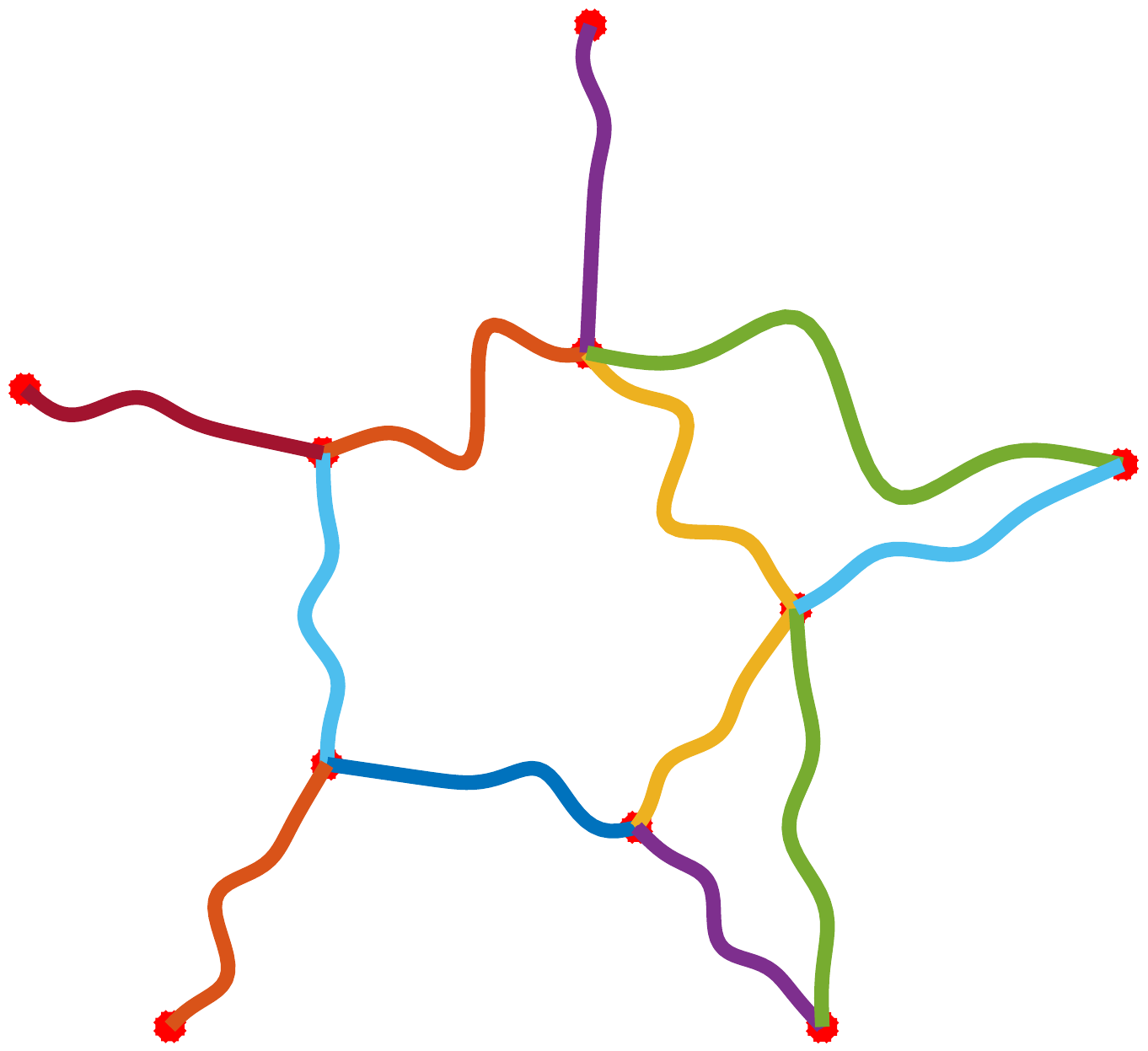} &
\includegraphics[height=0.7in]{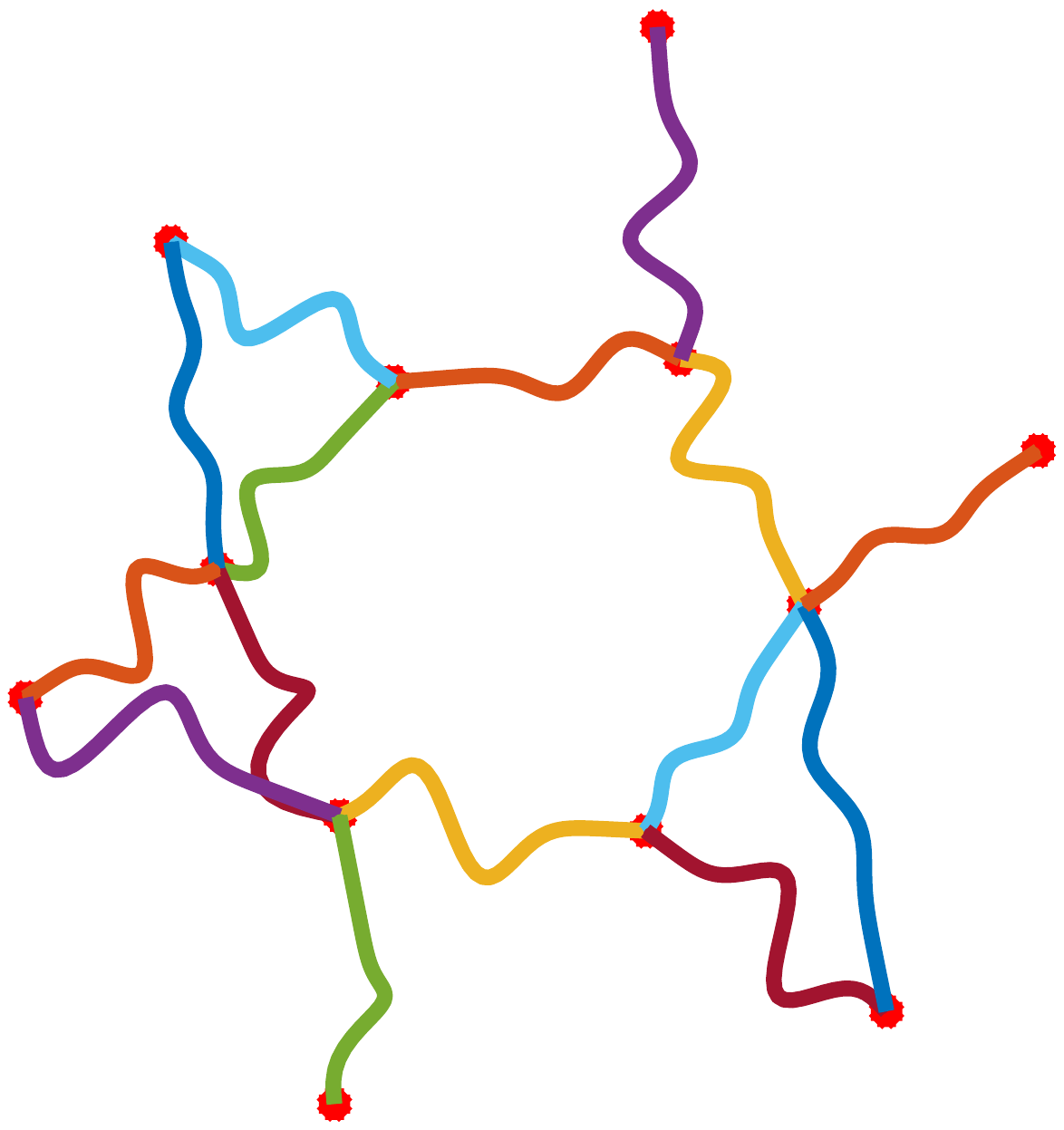} &
\includegraphics[height=0.7in]{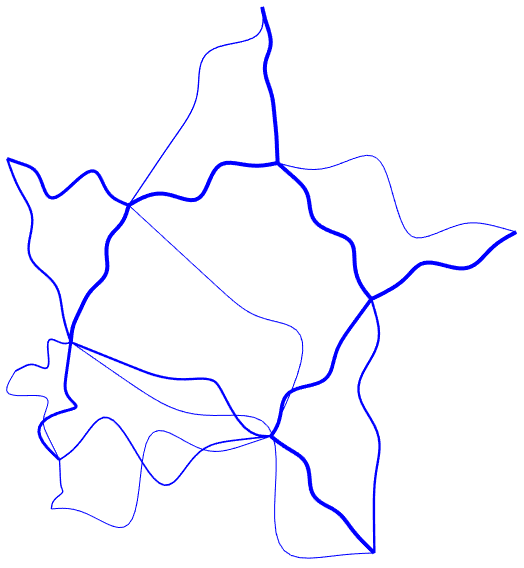}\\
 & & & & & & Method 2 \\
\hline
\multicolumn{7}{|c|}{\includegraphics[height=0.6in]{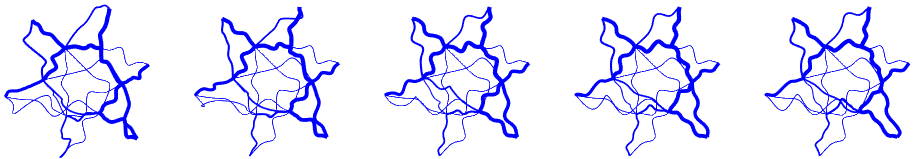}} \\
\multicolumn{7}{|c|}{First Principal Direction} \\
\hline
\multicolumn{7}{|c|}{\includegraphics[height=0.6in]{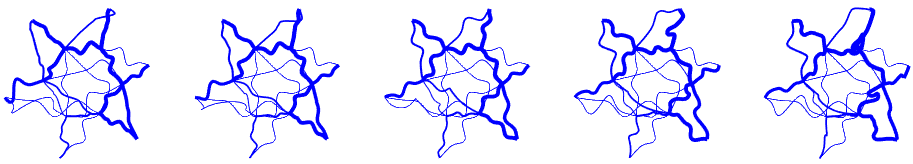}} \\
\multicolumn{7}{|c|}{Second Principal Direction} \\
\hline
\end{tabular}
\caption{Mean and PCA of 12 sample graphs.
The thickness of edges in the mean shape represents how often an edge is present in 
sample shapes. For PCA, in each row, the middle shape is the mean (method 1) while the right sides and left  sides are perturbation from mean by $\pm 1,\pm2$ square-root 
of the singular values.}
\label{fig:simu_mean_pca}
\end{center}
\end{figure}

We present an example of computing mean graphs in Fig. \ref{fig:simu_mean_pca}. 
The top left side shows a set of 12 graphs whose mean is being computed. While these
graphs have a common skeletal structure -- a ringlike interior with radial offshoots -- 
they also differ significantly in terms of the number of nodes/edges and shapes of edges. 
The mean of these graphs, computed using Algorithm \ref{algo:mean}, is shown in the figure's top right. It provides a reasonable visual and mathematical representation of the sample graphs, capturing the overall ringlike skeleton. The thickness of edges in the mean graph represents the frequency at which they appear in the given samples. One can always prune the thinner edges to improve clarity and focus on the larger, common structures. 
\\

\noindent {\bf Method 2 -- Sequential Approach}:  This approach starts with any two graphs and computes their mean to initialize the current mean. It then sequentially compares the current mean and one additional individual graph at each time and computes their pairwise weighted mean. This weighted mean becomes the new current mean for the next step. The full algorithm is given below.  The theoretical properties of this estimator of population 
mean are discussed in \cite{vemuri-AOAS:2019}.

\begin{algorithm}
\caption{Graph Mean in ${\cal G}$}
\label{algo:mean2}
\begin{flushleft}
Given adjacency matrices $A_{i}$, $i=1,..,m$:
\end{flushleft}
\begin{algorithmic}[1]
\State Find the constant speed geodesic between $A_1$ and $A_2$ and set $\mu$ to be the halfway point.
\State For each $i=3,4,\dots,m$, find the constant speed geodesic between $\mu$ and $A_i$. 
Set $\mu$ to the point at ${1 \over i}$th
distance from the previous $\mu$ along that new geodesic.  
\end{algorithmic}
\end{algorithm}

The top right panel of Fig.~\ref{fig:simu_mean_pca} also shows the result of computing the mean shape
using Algorithm~\ref{algo:mean2}. This shape denotes the mean of 12 graphs shown on the left side of this figure. Comparing this mean shape with the mean computed using Algorithm~\ref{algo:mean}, 
we see a lot of structural similarities, but we also see some visible
differences in the shapes of individual edges across the two means. Ideally, these two results should be identical. 
We attribute these differences to several factors, including the numerical errors involved in different steps of these procedures, especially 
Algorithm~\ref{algo:mean}. The differences can also result from each solution being 
a local rather than a global minimizer of the cost function.

\subsection{Tangent PCA in Graph Shape Space}
Graphical shape data is often high dimensional and complex, requiring tools for 
dimension reduction for analysis and modeling. 
In past shape analysis, the tangent PCA has been used to perform dimension reduction and discover dominant modes of variability in the shape data. Given the graph shape metric $d_g$ and the definition of shape mean $A_{\mu}$, we 
can extend TPCA to graphical shapes in a straightforward manner. 
As mentioned earlier, due to the non-registration of nodes in the raw data, the application of TPCA directly in $\mathcal{A}$ will not be appropriate. 
Instead, one can apply TPCA in the quotient space $\mathcal{G}$, as described in  Algorithm \ref{algo:PCA}. 
After TPCA, one can represent the graphs using low-dimensional Euclidean coefficients, which facilitates further statistical analysis. Besides TPCA, one may pursue geodesic PCA to reduce dimensions [\cite{bigot2017geodesic,calissano2020populations,cazelles2018geodesic,huckemann2010intrinsic}] ---implementation of geodesic PCA for elastic graphs will be a direction of future research.

\begin{algorithm}
\caption{Graph TPCA in $\mathcal{G}$}
\label{algo:PCA}
\begin{flushleft}
Given adjacency matrices $A_{i}$, $i=1,..,m$:
\end{flushleft}
\begin{algorithmic}[1]
\item Find the mean $A_{\mu}$ using Algorithm \ref{algo:mean} or \ref{algo:mean2}.
They results in the mean and the registered graphs $A_i^*, i = 1,2,..,m$.
\item For each $i$, evaluate the shooting vectors 
$v_i=(A_i^*-A_{\mu})$ as elements of $T_{A_{\mu}}(\mathcal{G})$ (the tangent space 
of ${\cal G}$ at $A_{\mu}$). 

\item   Perform PCA using the shooting vectors $\{v_i, i=1,2,\dots, m\}$ in $T_{A_{\mu}}(\mathcal{G})$.
Obtain principal directions and singular values for the principal components. 
\end{algorithmic}
\end{algorithm}

Bottom of Fig. \ref{fig:simu_mean_pca} shows an example of this TPCA procedure for graphical shapes. 
The figure shows shape variability
along the first two principal directions. 
As we can see, the first principal direction mainly
changes the smaller edges' shapes since the largest path is essentially the same across all the graphs.

\section{Brain Artery Networks}
Having developed tools for registering, comparing, and summarizing graphs using 
means and covariance, we now turn our attention to BAN data analysis~[\cite{bullitt2005vessel}]. 

We study the data from a geometric point of view and analyze these brain networks' shapes.
From an anatomical perspective, it seems natural to divide the full network into four components, as shown in  Fig. \ref{fig:example}. This division helps us focus on comparing individual components across subjects and makes the computational tasks more efficient.  
The original data has $98$ subjects, but we remove six that are difficult to separate into components, resulting in a sample size of $N=92$.
In the supplementary material~[\cite{supplement}], we provide some relevant statistics on the numbers of nodes and edges in the four components over the selected sample.
As these histograms show, these graphs differ significantly in the numbers of nodes and edges across subjects. 
Additionally, there are large differences in both the shapes and the patterns of arteries forming these networks. 
Consequently, analyzing the shapes of these BANs is quite challenging and remains relatively unexplored in the past.

\subsection{Geodesic Deformations}
\begin{figure}
\begin{tabular}{|c|@{}c@{}|@{}c@{}|}
\hline
Components & Full Geodesic & Pruned Geodesic \\
\hline
\raisebox{0.35in}{Left} &
\includegraphics[height=0.8in]{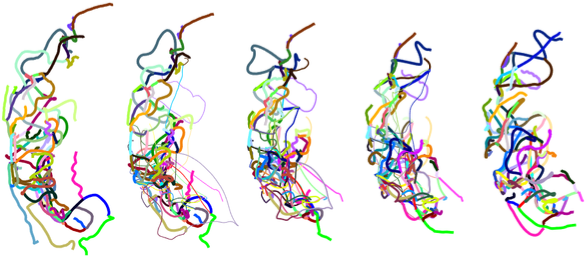}&
\includegraphics[height=0.8in]{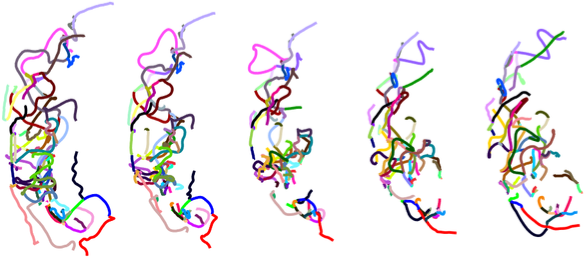} \\
\hline
\raisebox{0.35in}{Right} & 
\includegraphics[height=0.8in]{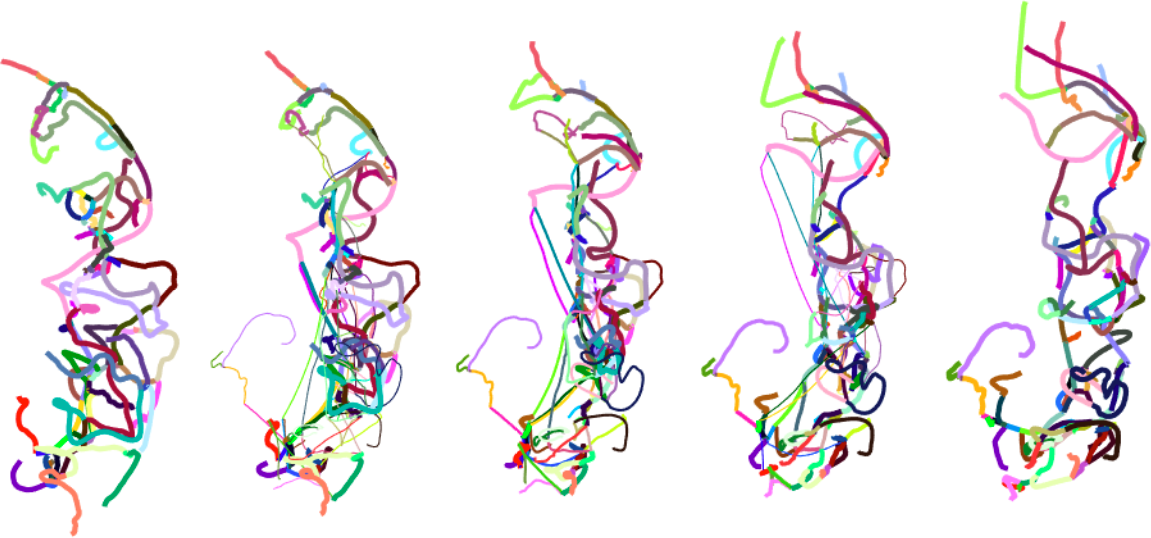}&
\includegraphics[height=0.8in]{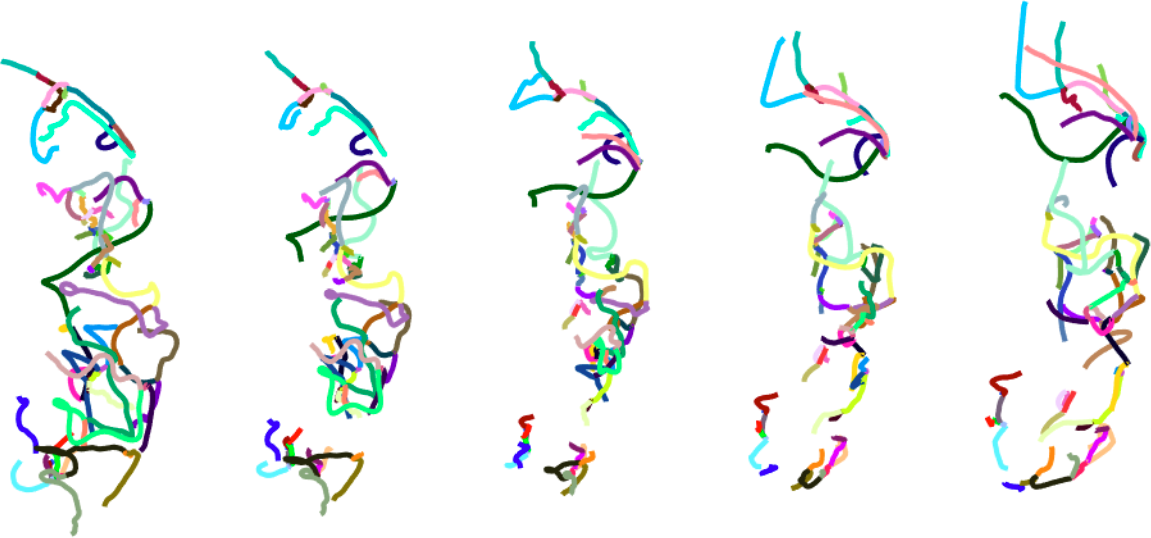}\\
\hline
\raisebox{0.1in}{Top} & 
\includegraphics[height=0.35in]{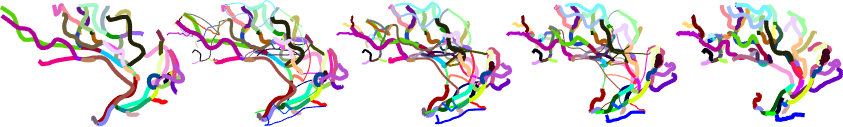}&
\includegraphics[height=0.35in]{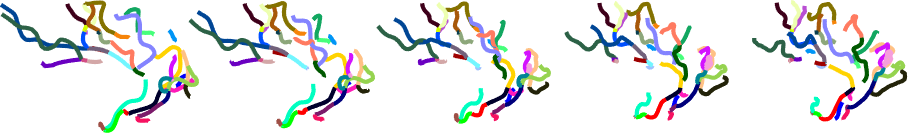}\\
\hline
\raisebox{0.1in}{Bottom} & 
\includegraphics[height=0.4in]{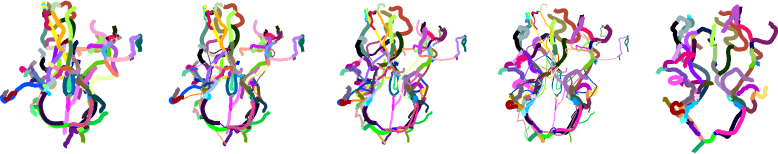}&
\includegraphics[height=0.4in]{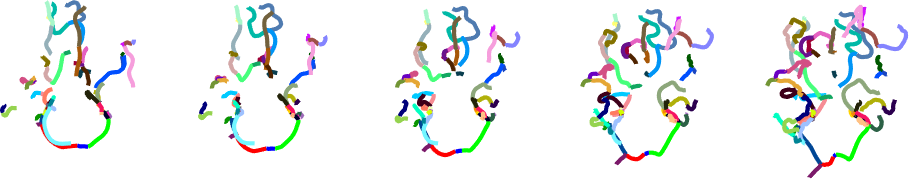}\\
\hline
\end{tabular}
\caption{Geodesics between BAN components. In each row we first show the full geodesic and then 
show a pruned geodesic where the unmatched edges are dropped.(GIF animations of these geodesics are provided with
the supplementary material~[\cite{supplement}].)}
\label{fig:brain_artery_geodesic}
\end{figure}

We use the techniques developed in this paper to compute geodesic paths between 
 BAN components and present some examples in Fig. \ref{fig:brain_artery_geodesic}. 
We use both the edge and node attributes in these experiments, with $\lambda$ selected manually through trial and error.
The first column in each row shows a geodesic comparing a BAN component of one subject
to that of another, as elements of $\mathcal{G}$.
To improve visual clarity, we remove some unmatched edges from the graphs and plot the same geodesic again in the right column.
We have used color-coding of edges to show registration and to track the deformation of each edge. These geodesics are useful in several ways. They provide registrations of arteries across networks, and they
help follow deformations of matched arteries from one network to another. 

\subsection{Average BAN Shapes}
\begin{figure}
\begin{center}
\includegraphics[width=0.45\linewidth]{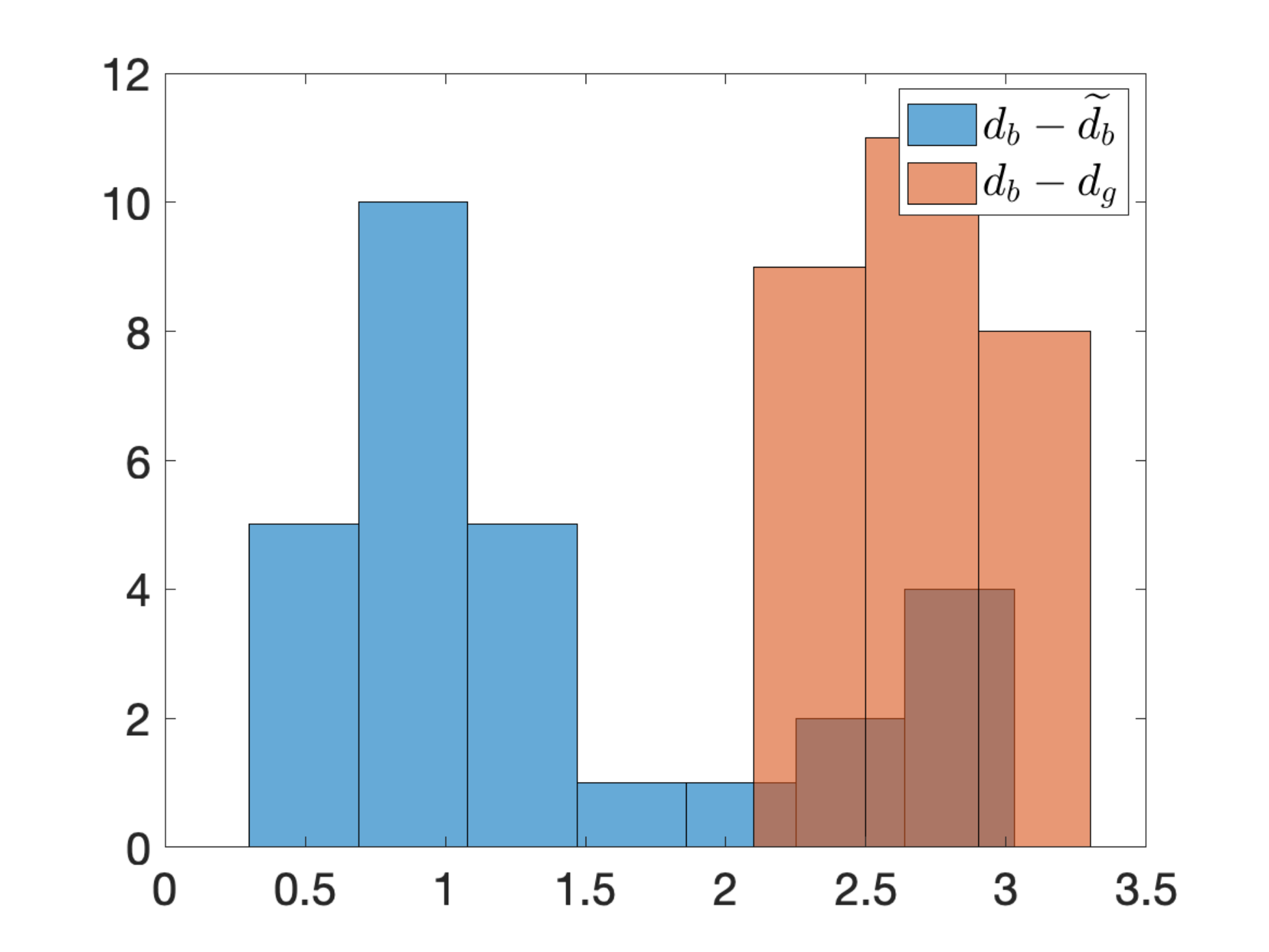}
\caption{Histograms of decreases in distances due to: (1) pairwise registration (red) and (2) registered to the largest graph (blue).}
\label{fig:hist}
\end{center}
\end{figure}

For the given $92$ BANs, it is informative to compute their mean shape. Since the computational cost of pairwise matching of graphs is high, the computation of mean
graph using Algorithm~\ref{algo:mean} becomes very expensive.  
To accelerate this process, we approximated the mean algorithm by registering each graph to the largest size graph in the dataset. We then used the fixed registration to compute the mean. 
This registration is not optimal, but it represents a tradeoff between accuracy and computational cost. 
In order to quantify this approximation, we perform a simple experiment. We take a small set of BANs, say 8 graphs, and compute 
three different pairwise distance matrices for them: (1) the distance $d_b$ without any registration, 
(2) the distance $\tilde{d}_b$, which is pairwise $d_b$ after registering all of them to the largest BAN in the group, and (3) 
the distance $d_g$ after optimal pairwise registration. Fig.~\ref{fig:hist} shows the results using histograms of two quantities: 
(1) $d_b - \tilde{d}_b$ in blue, and (2) $d_b - d_g$ in red. This plot underscores that the proposed approximation provides reasonable gains in distance computations at very small computational cost. 
Here we have tried one solution, {\i.e.}, matching all the graphs to the largest
graph, but there are other choices too, each representing its own tradeoff. 
For instance, one can cluster the graphs into smaller groups and then iteratively improve the registration between and across the clusters. Another idea is to train graph neural networks for performing graph matching. In case there is sufficient training data available, neural networks can provide very efficient computational solutions. Of course, any such solution can be good or bad depending upon actual structural variability in the data. One can always return to the pairwise optimal matching if accuracy is the real concern and the computational cost is less relevant.

\begin{figure}
\begin{center}
\includegraphics[height=2in]{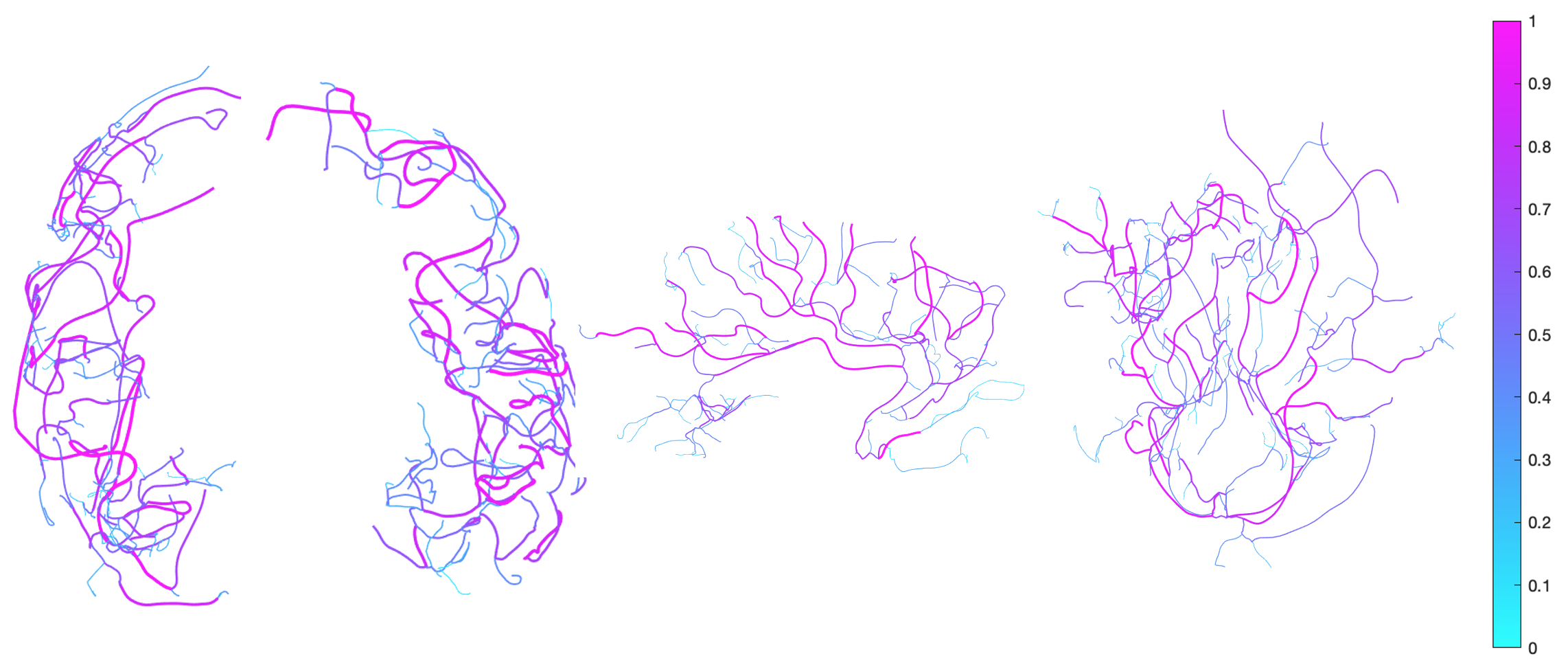}
\caption{Average shapes of BAN components across 92 subjects. From left to right, it is the mean shape for left, right, top, and bottom components, respectively. The color and thickness of an edge represents its proportional presence in individual subjects --
a more red and thicker edge indicates it is present in more subjects. }
\label{fig:brain_artery_mean}
\end{center}
\end{figure}

Figure \ref{fig:brain_artery_mean} shows the
resulting mean shapes for each of the four components across $92$ subjects.
Since these $92$ graphs differ in the number, connectivities, and shapes of edges, it is not easy to visualize and interpret the mean shapes. One needs to view 3D displays of these shapes to appreciate how well these means capture the common structures in individual graphs. 
We use the color and thickness of edges to denote the proportion of individual graphs in which that particular edge is 
present. 
As expected, these mean shapes show a smoother, 
broader representation of individual shapes and mainly preserve 
connectivity patterns present in the data. 

The computation of an average shape is a significant, novel result and has not been achieved previously for BANs
or any similar graph data. Its importance lies in our need to separate structures 
common to all subjects from structures that distinguish subjects from each other. By separating this
variability, one can focus on individual differences and model this variability using statistical models. 
Once the common structures have been removed, the subsequent statistical analysis simplifies greatly. 
We present these studies in the next few sections. 

\subsection{PCA-Based Analysis of Covariate Effects on Shape}
An essential use of the proposed framework is in understanding  
the effects of covariates, such as gender and age, on shapes of BANs.  
Due to the high dimensionality and complex nature of BANs, this task is nearly impossible in its original space. We use the elastic shape analysis framework to compute average shapes, extract individual variability, and focus on modeling this individual variability against covariates. As earlier, we do this component-wise, {\it i.e.}, a separate analysis of each component of BAN.

We first perform Graph PCA using Algorithm \ref{algo:PCA} on BAN components and represent each subject data into a low-dimensional vector space via projection.  
A BAN component with 197 nodes and 50 sample points along each edge has a 
discrete representation with $3 \times 50 \times 197 \times 197 = 5821350$ elements.
However, using Graph PCA, we can retain $80 \% $ of the variability
in the original data using only $60$ principal components.
To avoid the confounding effect of artery size, we rescale edges by the total artery length for each graph. 
As a result, we can focus on gender and age effects on the shapes of BANs.

\subsubsection{Gender Effect on BAN Shapes}
To study the effect of gender on the arterial graph shapes, 
we implement a two-sample $t$-test on the first principal scores and a Hotelling's $T$-squared test on the first few principal scores. 
The resulting $p$-values can be found in Table \ref{tab:brain_artery_gender}.
Most of the $p$-values are high; thus, we do not find any significant difference between the BAN shapes and the gender. To further investigate this result, 
we also applied a
permutation test (described in the following paragraphs) and
obtained similar results; see Table \ref{tab:test_perm}.
We note that~\cite{bendich2016persistent} and~\cite{shen2014functional} reported multiple insignificant and 
significant $p$-values on the same data using very different mathematical representations from ours.
Whether there is an anatomical shape difference between BANs of female and male 
subjects remains an open question and requires further investigation.

\begin{table}
\caption{Testing of gender effect on principal scores of shapes of brain arterial networks.}
\centering
\begin{tabular}{|c|c|cccc|} 
 \hline
         & t test & \multicolumn{4}{c|}{Hotelling's T-squared test }  \\ 
         & PC-1 &  PC 1-2  & PC 1-3 & PC 1-4 & PC 1-5  \\
 \hline
 \hline
 Left  & 0.1354 & 0.5228  & 0.4333 & 0.5074 & 0.4009\\
 \hline
 Right & 0.8868 & 0.0785  & 0.1630 & 0.0792 & 0.1210  \\
  \hline
 Top & 0.9236 & 0.6788  & 0.0676 & 0.1200 & 0.1400 \\
  \hline
 Bottom & 0.4005 & 0.0599 &  0.1256 & 0.1447 & 0.2328  \\
 \hline
\end{tabular}
\label{tab:brain_artery_gender}
\end{table}

\subsubsection{Age Effects on BAN Shapes}
\begin{figure}
\centering
\begin{tabular}{cccc}
\hspace{-0.5cm} \includegraphics[width=0.25\linewidth]{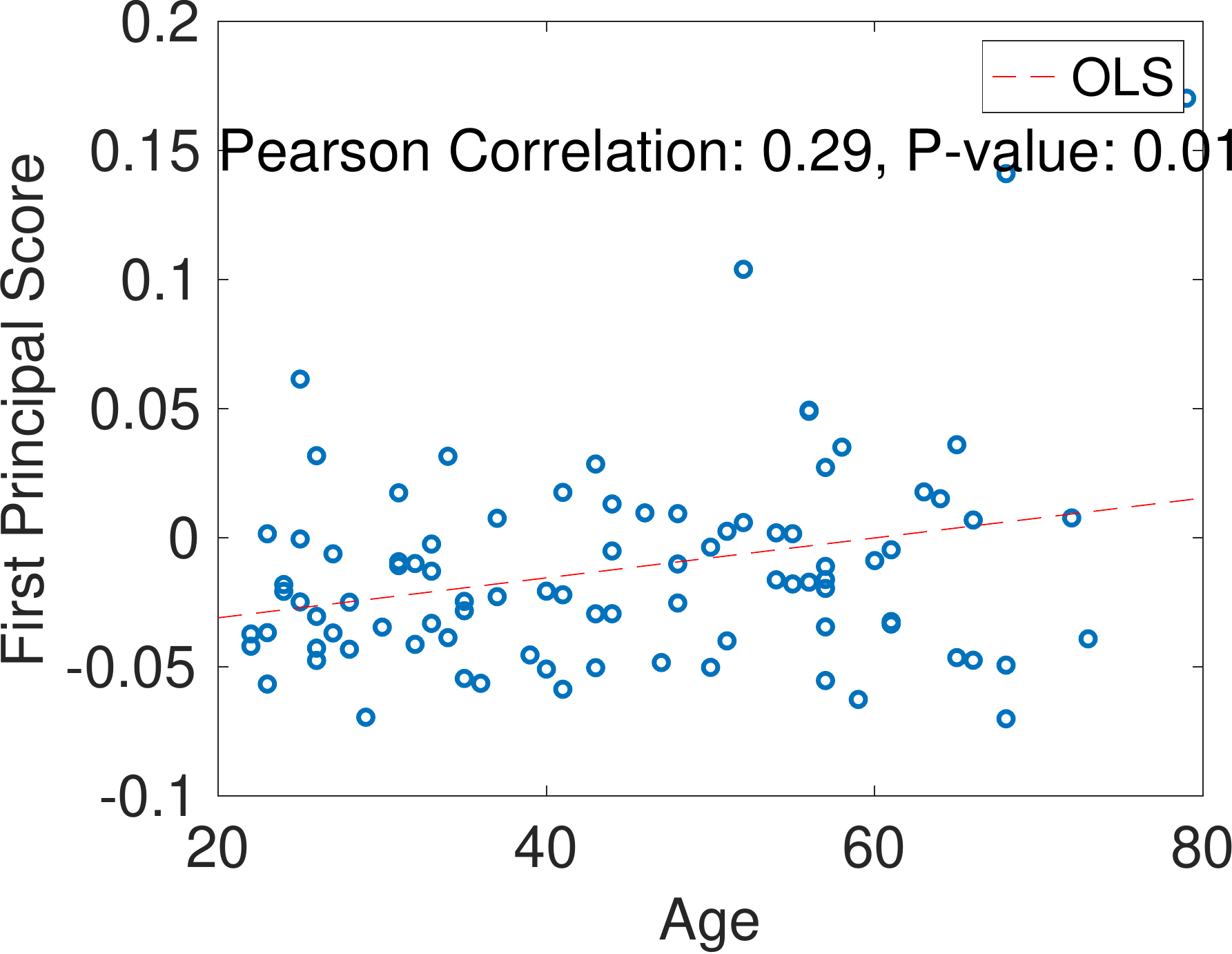} &
\hspace{-0.5cm} \includegraphics[width=0.25\linewidth]{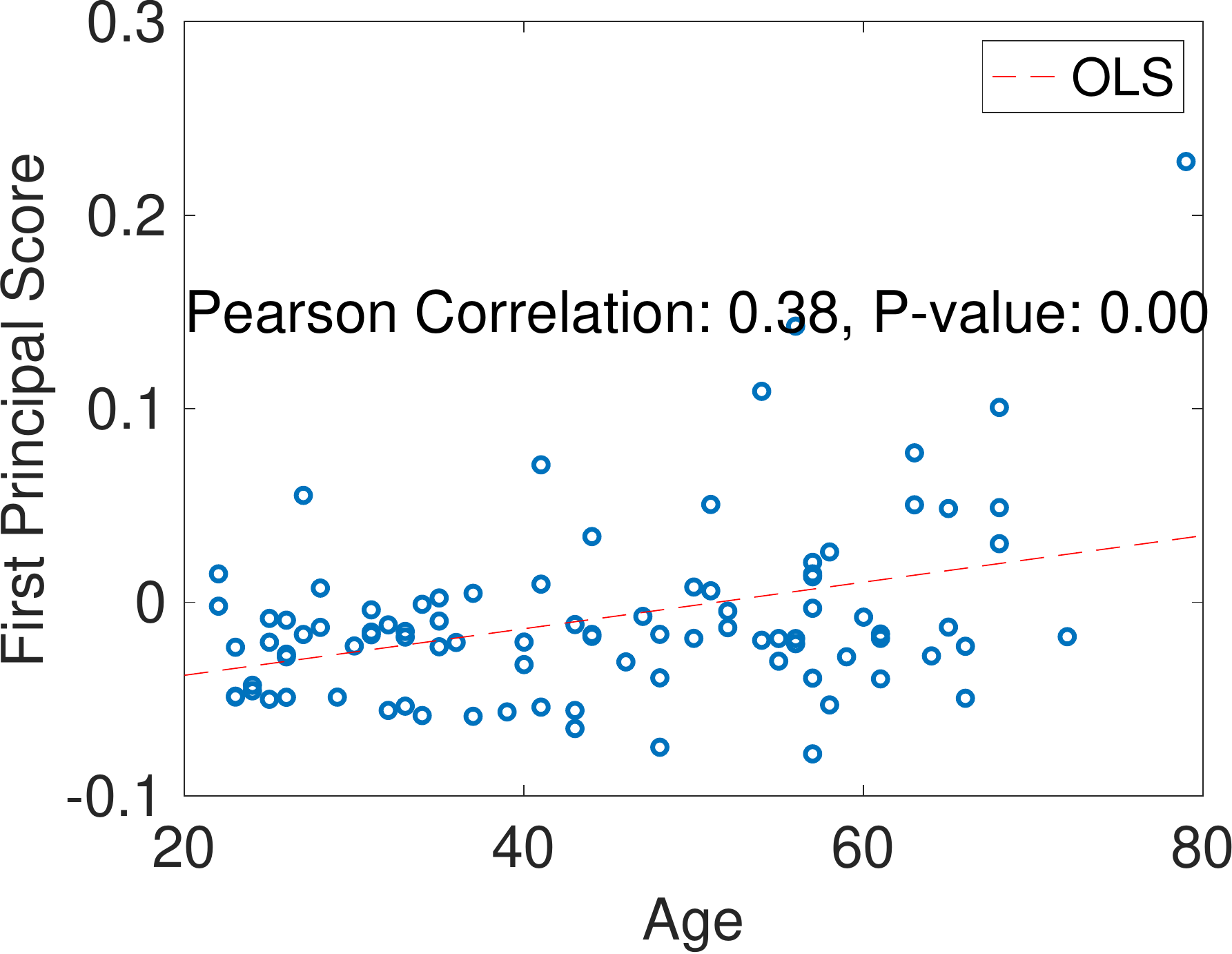} &
\hspace{-0.5cm} \includegraphics[width=0.25\linewidth]{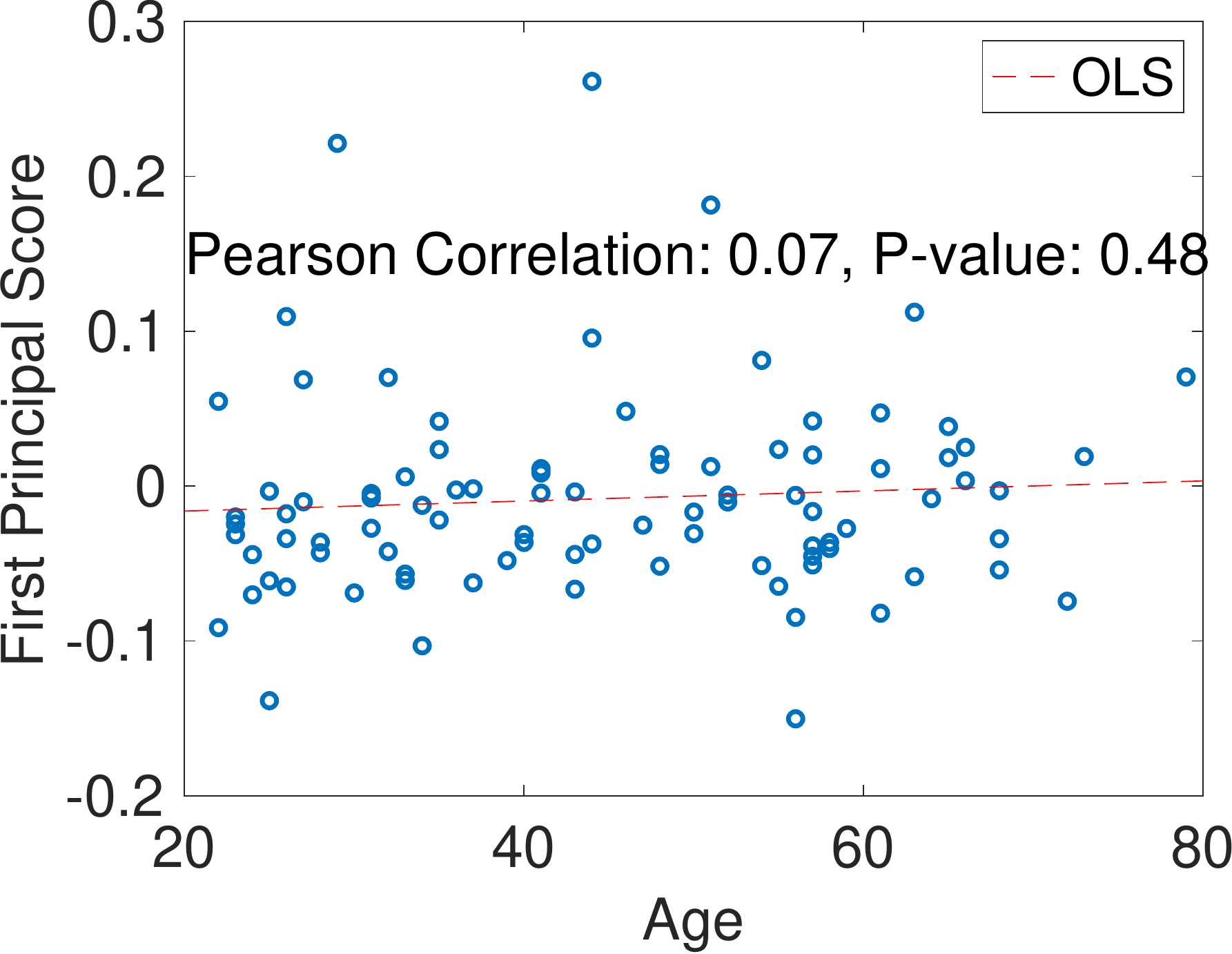} &
\hspace{-0.5cm} \includegraphics[width=0.25\linewidth]{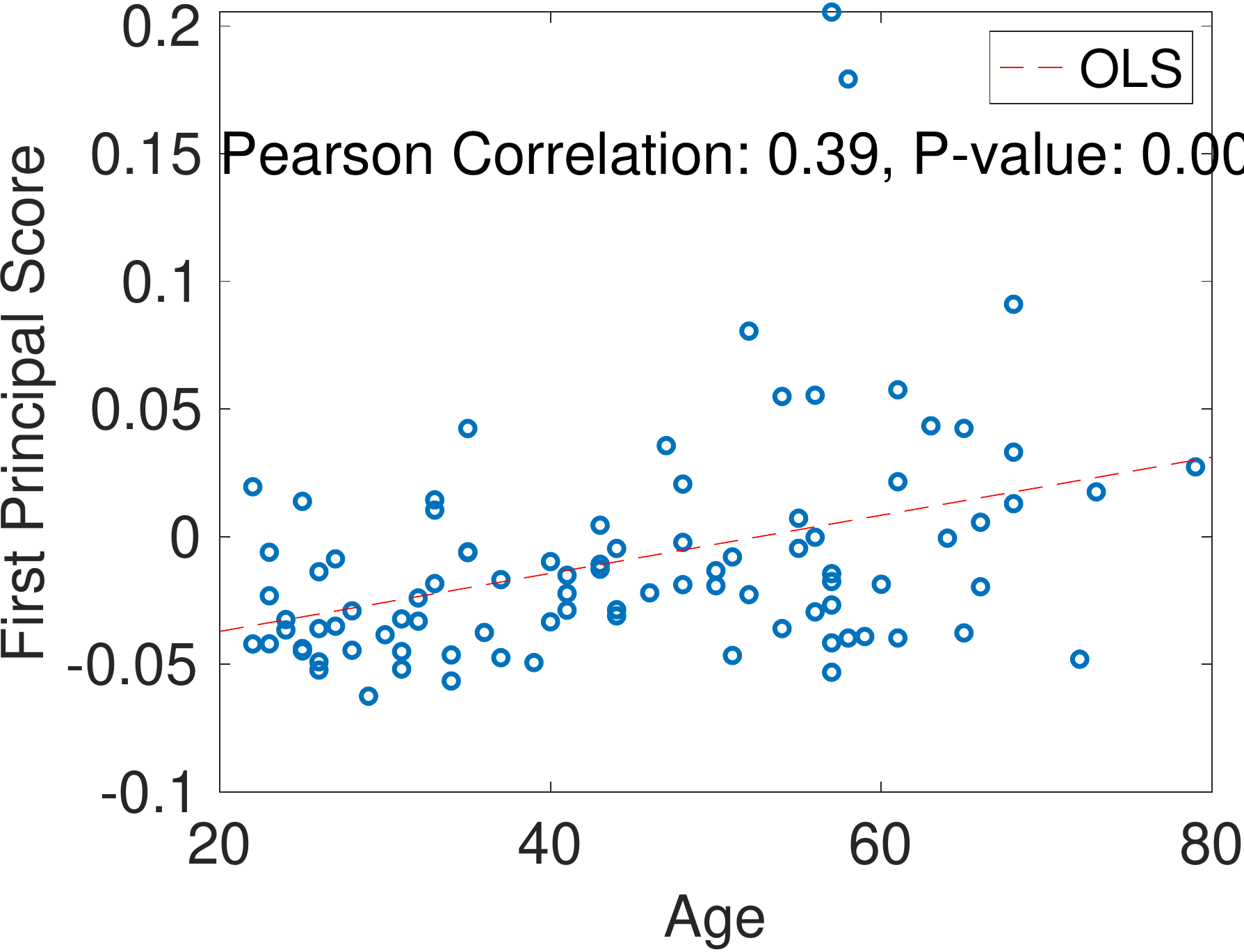}\\
Left & Right & Top & Bottom
\end{tabular}
\caption{Correlation between the age and the first principal score of shape of arterial networks. To better show the correlation patterns, we removed the maximum principal score for those four components, respectively. In addition, we also removed the second maximum for the left component.}
\label{fig:brain_artery_age}
\end{figure}

To study the effect of age of a subject on his/her BAN shape, we studied correlations between the
age and the PCA scores of arterial shapes. The results are shown in Fig. \ref{fig:brain_artery_age}. 
We found a significant linear correlation between age and first principal scores of brain arteries in most cases (in all except the top component). 
The correlation coefficients between the first principal shape score for the left, right, and bottom components and age are $0.29$, $0.38$, and $0.39$, respectively.
All of them are statistically significant, with corresponding $p$-values being very close to zero.
This result is similar to some published results in the literature but obtained using different mathematical representations than ours~[\cite{bendich2016persistent,shen2014functional}].
The strength of our approach lies in our ability to visualize the nature of deformations resulting from aging. 
As mentioned before, we also use a permutation test to validate the age effects.

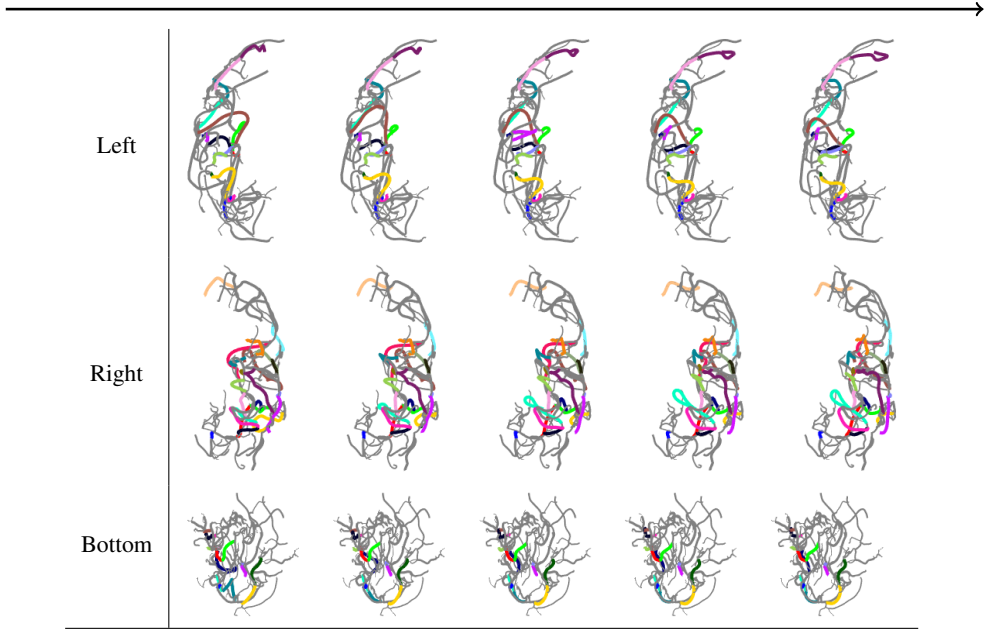
\begin{figure}
\begin{center}
{\bf \Large Age (from 22 to 79 years)}\\
\end{center}
\begin{tikzpicture}
\draw[->,line width=1pt] (5,0) to (18,0);
\end{tikzpicture}
\begin{center}
\begin{tabular}{c|l}
\raisebox{0.5in}{Left} & \includegraphics[height=1.15in]{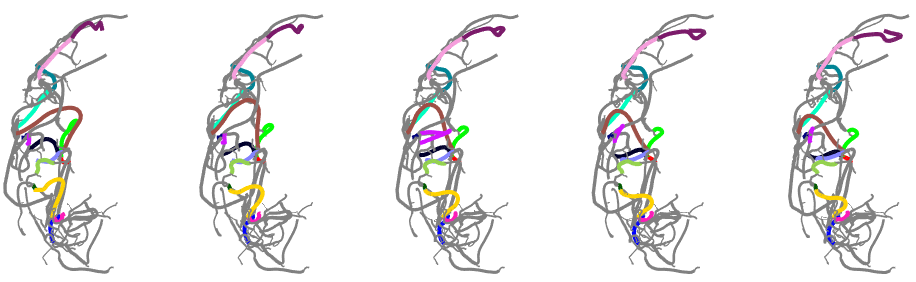}\\
\raisebox{0.5in}{Right} & \includegraphics[height=1.15in]{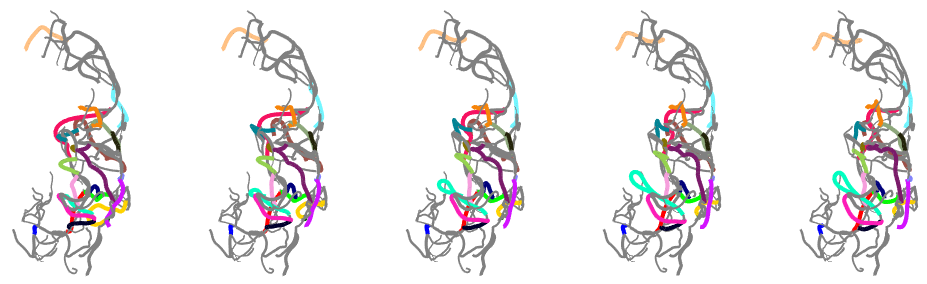}\\
\raisebox{0.35in}{Bottom} & \includegraphics[height=0.7in]{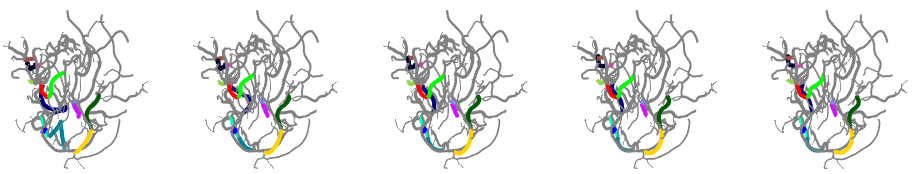}\\
\hline
\end{tabular}
\caption{Aging effect on brain arteries. For each component, from left to right is the mean shape deformation 
when going from age 22 to 79. The colors highlights edges that have large deformations.}
\label{fig:brain_artery_aging}
\end{center}
\end{figure}

Next, we pose an important question: How do the BAN shapes change as a subject gets older? 
The tools developed in this paper can be used to visualize the effects of aging on the shape of brain arteries, 
while feature-based methods proposed in the past cannot address this question. 
Using Algorithm \ref{algo:PCA}, we compute PCA scores for each BAN shape.
Treating these scores as a response and age as a predictor, we fit a zero-mean regression model to the data. 
Note that since principal scores can be used to reconstruct original graphs, we have a way 
of mapping these representations back to graphs and facilitate visualizations. 
Therefore, we can visualize the effects of aging on the shapes of BANs,
as in Fig \ref{fig:brain_artery_aging}. 
Here we fit a linear model using age to predict the first principal scores. 
To focus visualization on where the changes are occurring, 
we calculate the edgewise shape differences using the mean shape as baseline and use different
colors (rather than gray color) to denote edges with large deformations, {\it i.e.}, deformations that are larger than $50\%$ of the differences.

\subsection{Metric-Based Study of Covariates Effects on Shapes}
\begin{figure}
\centering
\begin{tabular}{@{}c@{} @{}c@{} @{}c@{} @{}c@{}}
\includegraphics[height=1.0in]{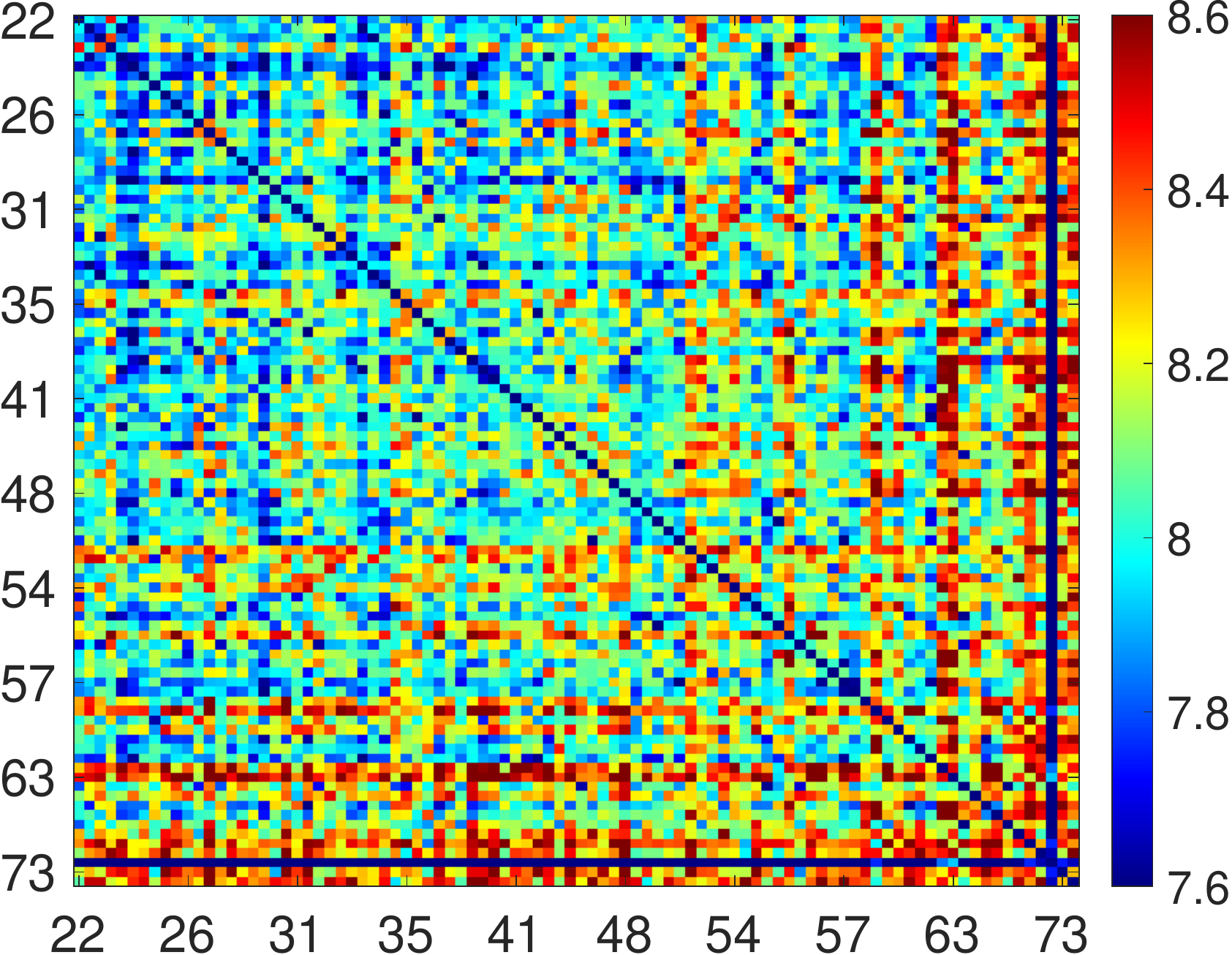} &
\includegraphics[height=1.0in]{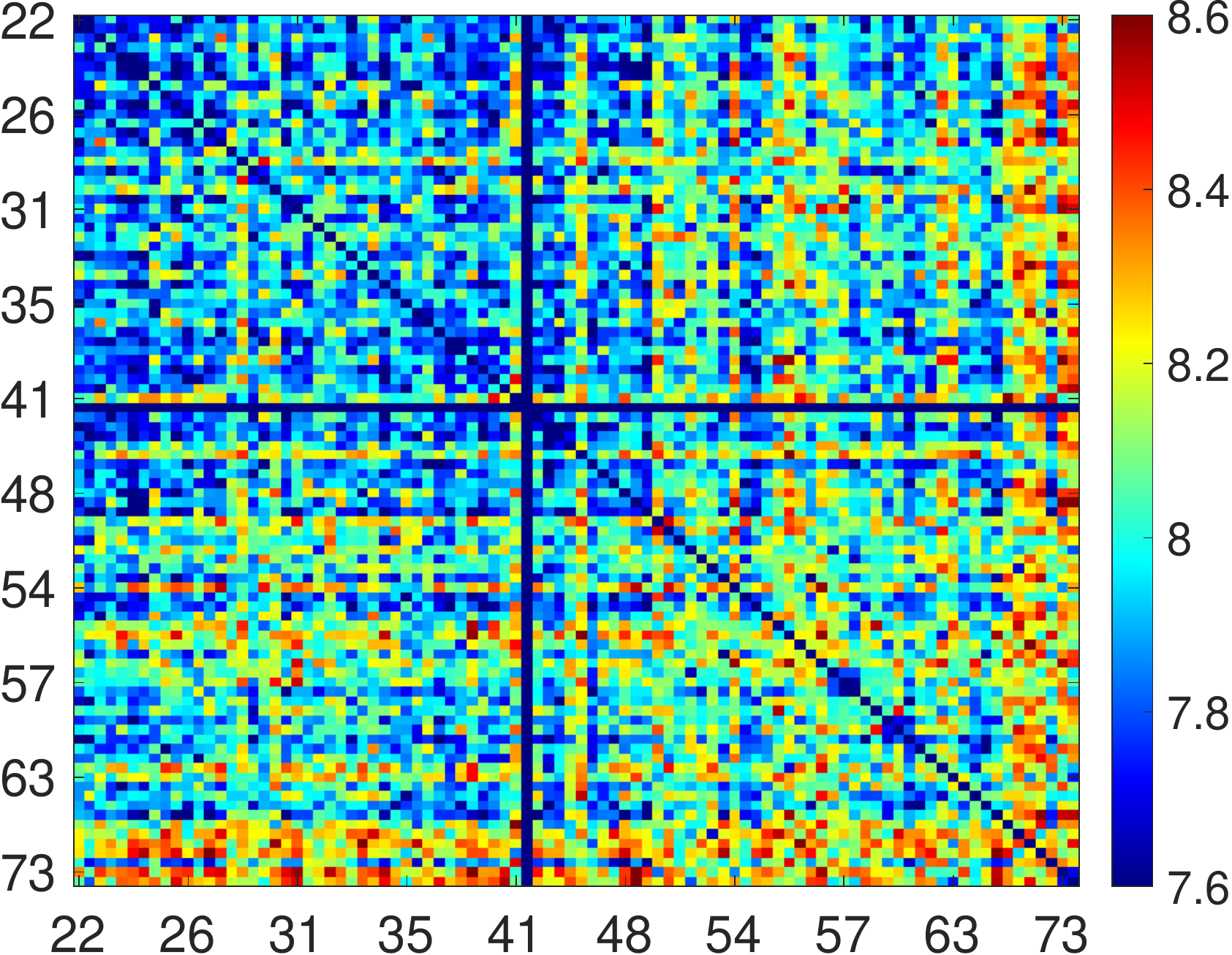} &
\includegraphics[height=1.0in]{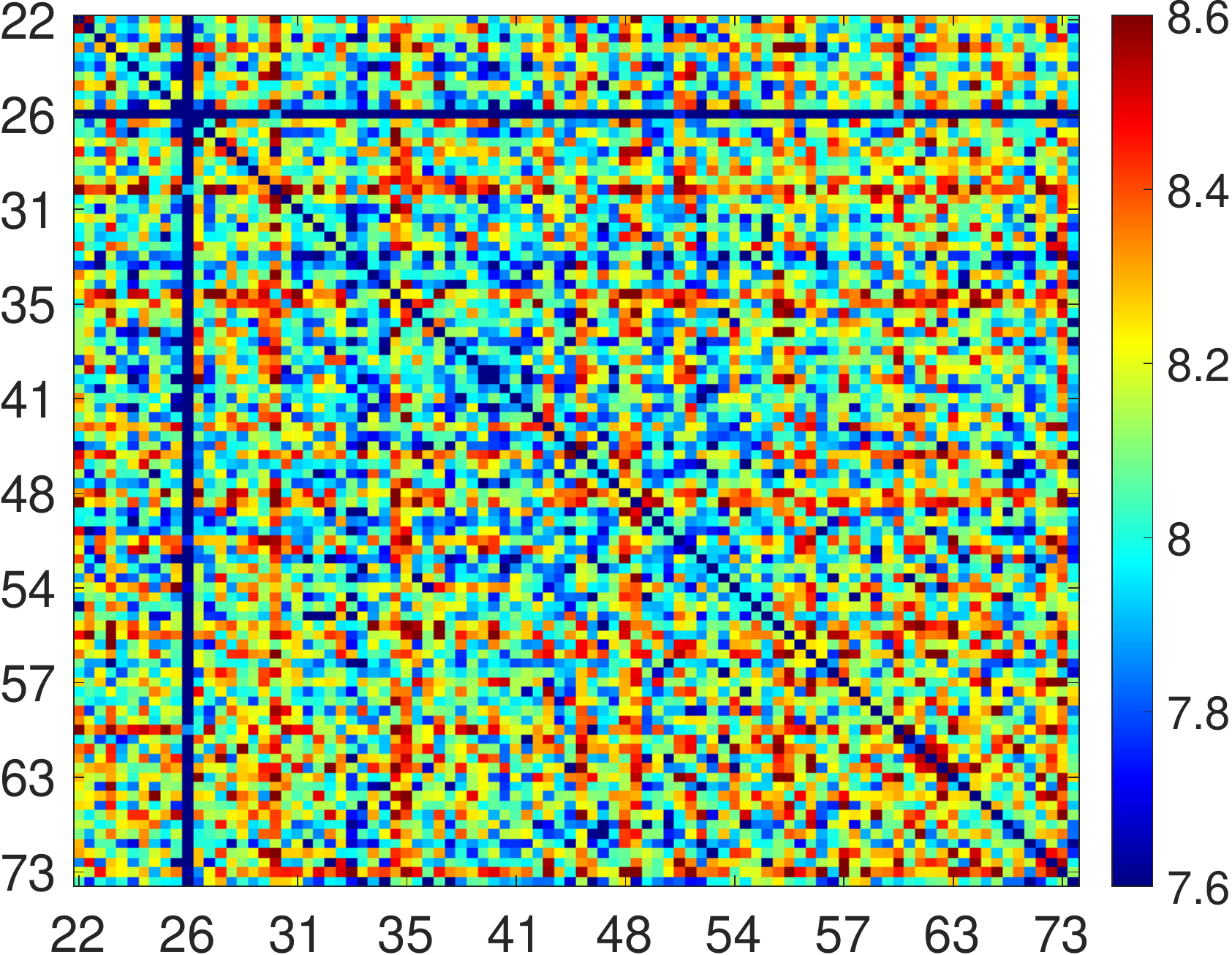} &
\includegraphics[height=1.0in]{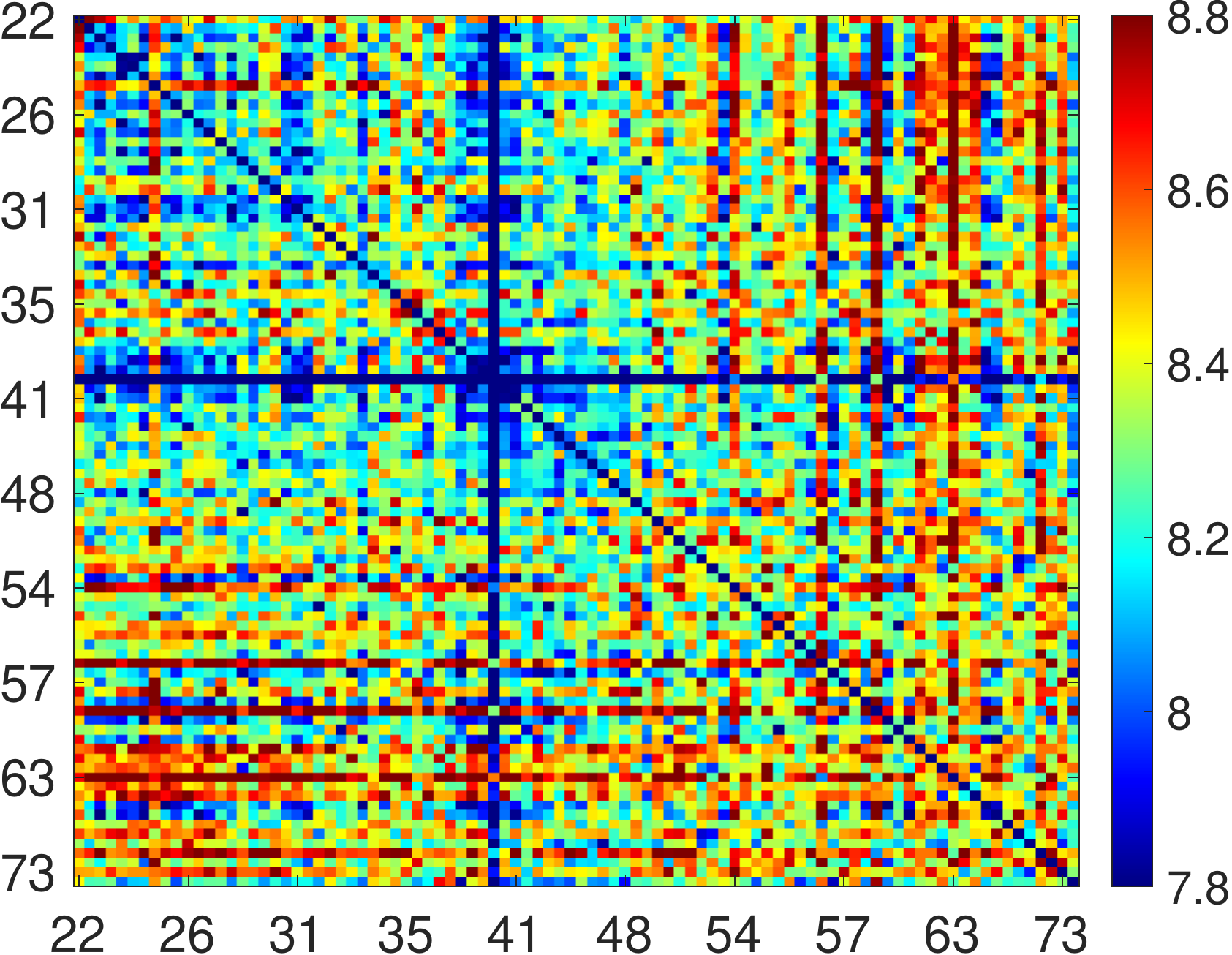}\\
Left & Right & Top & Bottom
\end{tabular}
\caption{Matrices of shape distances between BAN components of 92 subjects. 
The axes are labeled by ages.}
\label{fig:brain_artery_distmat}
\end{figure}

We also investigate the effects of covariates on full BAN shapes using the shape metric $d_g$ (Eqn. \ref{eq:metric}) directly. 
Fig. \ref{fig:brain_artery_distmat} shows matrices of pairwise distances between subjects, 
one matrix for each of the four components separately. 
(As mentioned before, we have scaled the edges by the total artery length; thus, the distances quantify only shape differences.)
We reorder the distance matrices by the age of subjects to help elucidate the effect of aging on shape variability. The color pattern of pixels in these matrices shows that shape distances
increase with age (darker red colors are towards the bottom right). 
This pattern implies that brain arterial networks' shape variability grows with age for three of the four components.
This pattern does not seem to hold for the top component. 

To further validate the gender and age effects, we implement a permutation test~ [\cite{hagwood-etal:2013}] 
based on shape metric $d_g$.
The basic idea is as follows. 
We divide the subjects into two groups -- older than 50 and younger than 50 -- 
and compute a two-sample shape test statistic specified in [\cite{hagwood-etal:2013}]. 
For two groups, labeled A and B, the test statistics is defined to be: 
{\small $$
 \left( {2 \over m_1m_2} \sum_{a_i \in A} \sum_{b_j \in B} d_g(g_{a_i}, g_{b_j}) - 
{1 \over m_1^2} \sum_{a_i \in A} \sum_{a_j \in A} d_g(g_{a_i}, g_{a_j})
- {1 \over m_2^2} \sum_{b_i \in B} \sum_{b_j \in B} d_g(g_{b_i}, g_{b_j}) \right)\ .
$$
}
Here $m_1$ and $m_2$ are the sizes of the two sets $A$ and $B$, respectively. 
We evaluate the significance of this value using a Permutation test.

That is, we repeat this process $30K$ times, 
each time randomly assigning subjects into different groups 
and computing the test statistics. 
Using a histogram of the $30K$ test statistics, we can compute the $p$-value of the real-data test statistic.
The result can be found in Table \ref{tab:test_perm}.
While the gender effect remains unclear, one can see a significant age effect on the brain arteries 
for the left, right, and bottom components. 
We also see that the top component remains relatively unchanged between young and older people. 
(We remind the reader that the pairwise distance used here is an approximation because of 
the considerable computational cost associated with the exact computation. 
Here we first match each graph to the largest graph in the dataset and compute pairwise shape distances 
between them without any 
further matching.)

\begin{table}
\caption{\scriptsize Permutation test of gender and age effect on distances of shapes of brain arterial networks.}
\centering
\begin{tabular}{|c|cccc|} 
 \hline
   &  Left  & Right & Top & Bottom  \\
 \hline
 \hline
 Gender  & 0.0385 & 0.2128  & 0.4835 & 0.0706\\
 \hline
 Age & 0.0006 & 0  & 0.4772 & 0  \\
 \hline
\end{tabular}
\label{tab:test_perm}
\end{table}

\section{Improving Registration Using Landmarks}
\label{sec:landmarkgraphmatching}
As stated earlier, matching parts of BANs is the most crucial bottleneck in the shape analysis of elastic graphs. The procedure laid out so far for graph matching is fully 
automated but, in the end, does not guarantee a global solution. 
One can potentially improve this solution in case there is some extra matching information. 
This knowledge can be in the form of some prominent points, called {\it landmarks}, that 
have known registrations across graphs. 
In this context, two issues arise: (1) How can we obtain the landmarks? and (2) How to incorporate
this extra information in improving dense registration of graphs? 

For the first issue, there are several possibilities. The landmarks may be provided by the domain experts using manual analysis. Another possibility is to perform a preliminary investigation of the data and extract some landmarks of interest. To facilitate this approach, one can use traditional graph-theoretic tools to discover and extract some prominent landmarks automatically. Examples of relevant tools include multi-resolution representations of graphs, clustering of nodes using graph spectra, 
and Dijkstra's method for finding the longest paths in a graph. For instance, one can use the longest arteries in 
BANs, extract some 
prominent nodes lying on them, and use them as landmarks for matching across BANs.

The second issue -- how to incorporate given landmarks registration in solving 
Eqn. \ref{eq:metric} -- is more methodological. We will assume that all landmarks are 
{\it nodes} in the original graphs, although one can relax this assumption albeit at the cost of some increased computational complexity. 
For these {\it registered nodes}, the corresponding entries of $P \in {\cal P}$ are fixed and no longer
part of the search. Assuming $n_0 \leq n$ to be the number of landmarks, the search
space is now reduced to $(n-n_0) \times (n-n_0)$ permutation matrices. 
This constrained search has also been called {\it seeded graph matching}
in the literature~[\cite{fishkind2012seeded}].

\begin{figure}
\begin{center}
\begin{tabular}{|@{}c@{}|@{}c@{} |@{}c@{}|}
\hline
\includegraphics[height=1.0in]{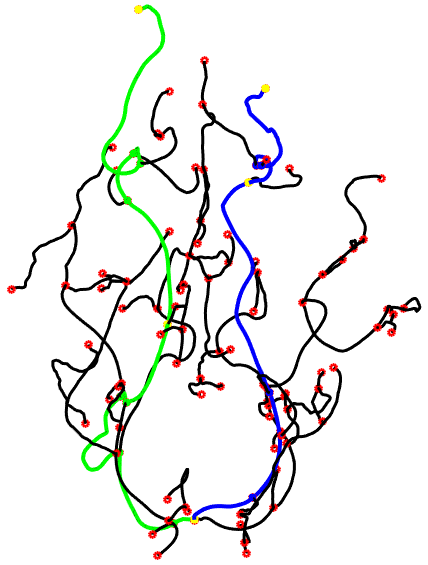} &
\includegraphics[height=1.0in]{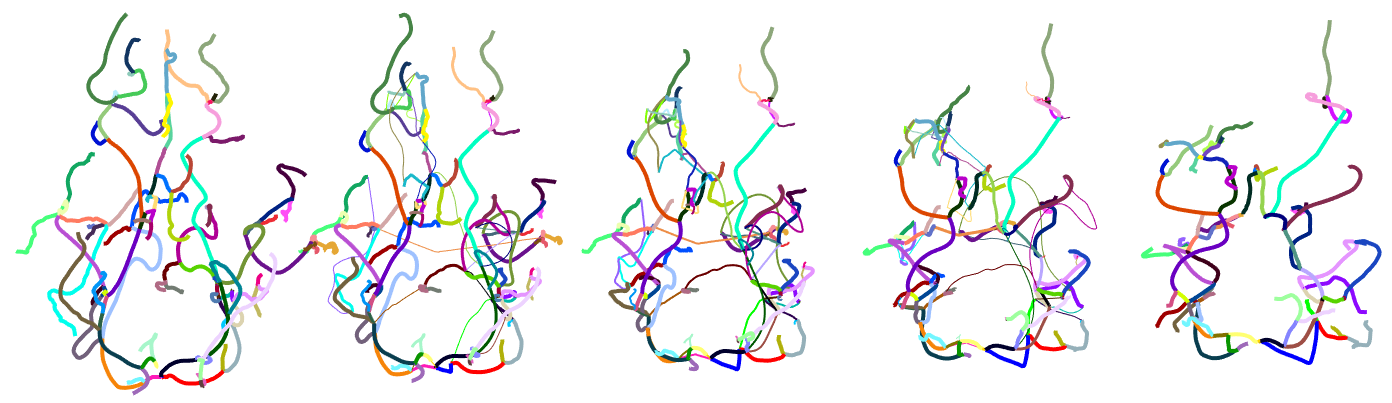}&
\includegraphics[height=1.0in]{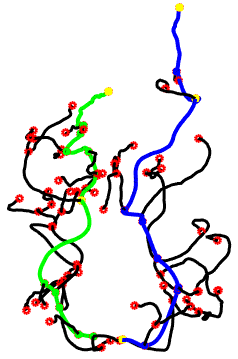} \\
Graph 1 & Full geodesic from graph 1 to graph 2 & Graph 2 \\
\hline
& \includegraphics[height=1.0in]{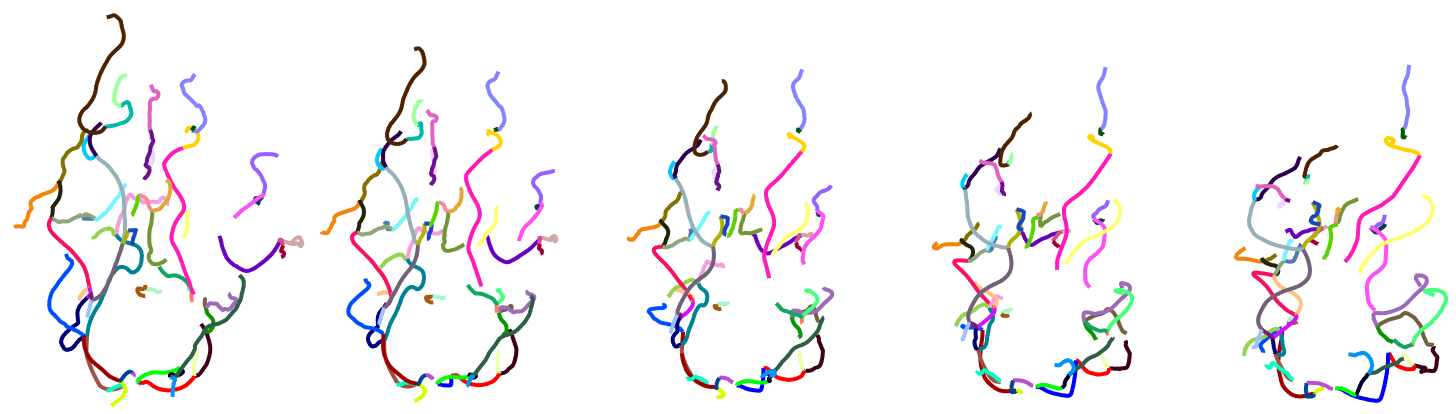}  &\\
 & Geodesic with unmatched branches pruned & \\
\hline
\end{tabular}
\caption{Geodesic path between two BANs (bottom components) using five registered landmarks.}
\label{fig:geodesic-landmark}
\end{center}
\end{figure}

\begin{figure}
\begin{center}
\begin{tabular}{|@{}c@{}|@{}c@{} |@{}c@{}| @{}c@{}| @{}c@{}| @{}c@{}|}
\hline
\includegraphics[height=1.0in]{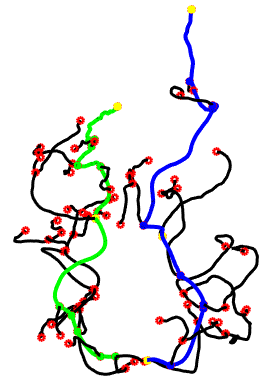} &
\includegraphics[height=1.0in]{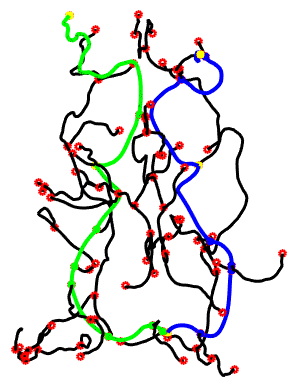} &
\includegraphics[height=1.0in]{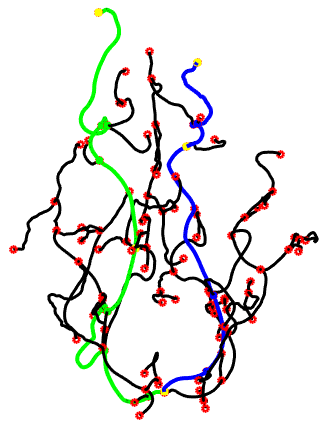} &
\includegraphics[height=1.0in]{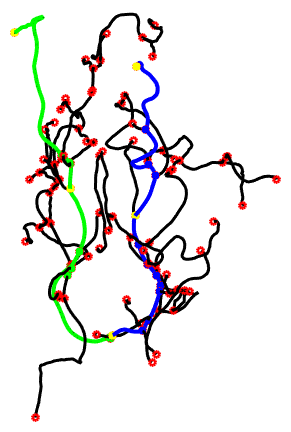} &
\includegraphics[height=1.0in]{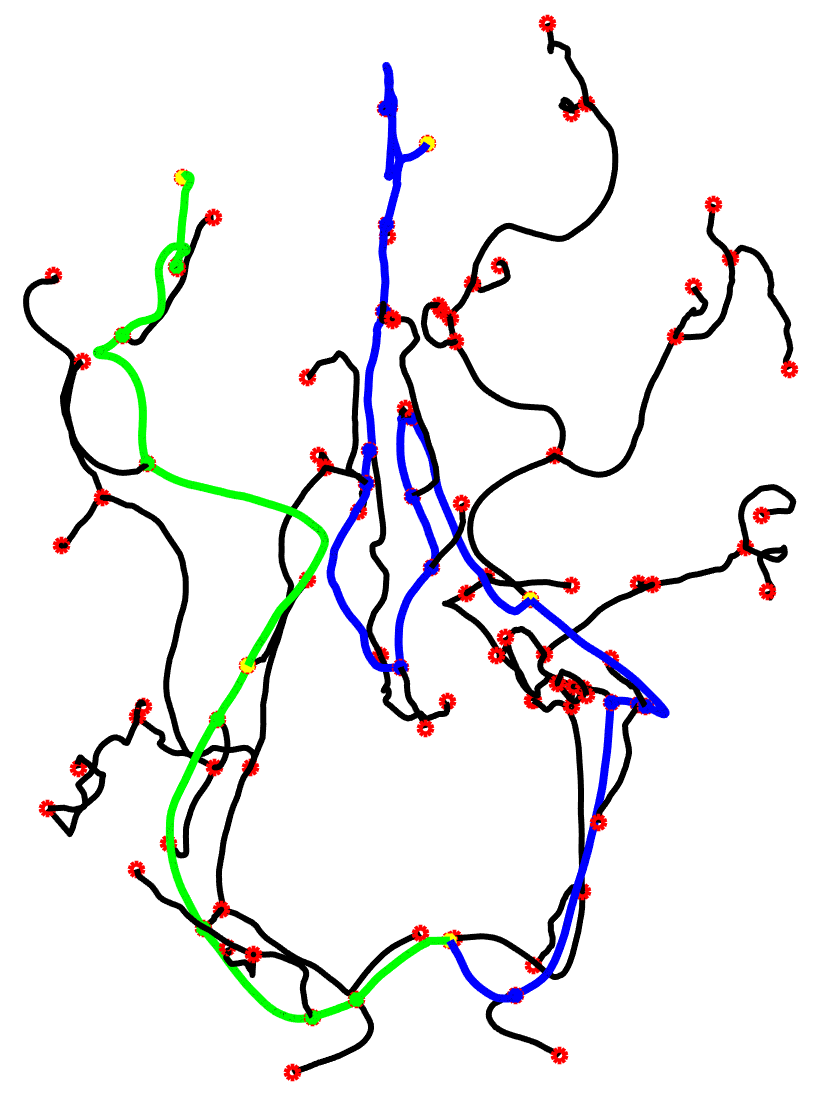} &
\includegraphics[height=1.0in]{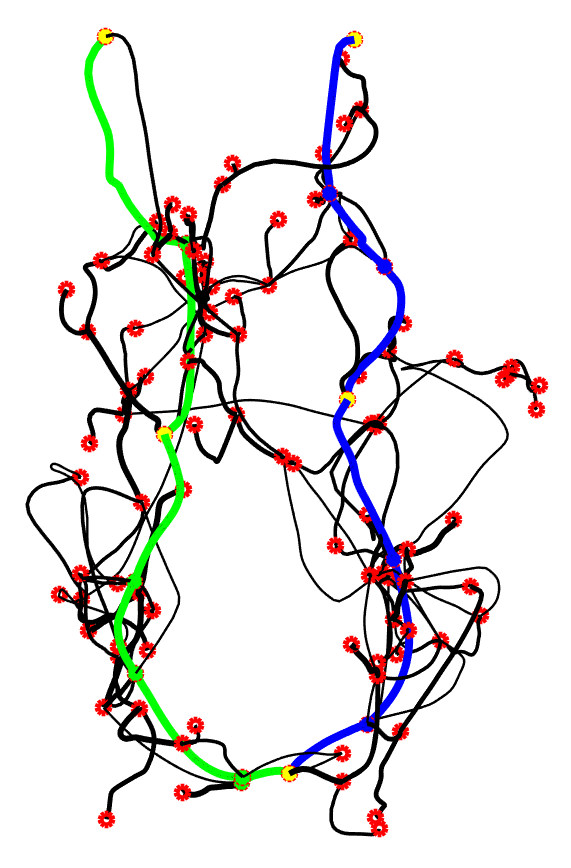} \\
\hline
\multicolumn{5}{|c|}{BAN-bottom for five subjects} & Mean Shape\\
\hline
\end{tabular}
\caption{Sample mean of BAN components using improved landmark-based registration.}
\label{fig:mean-landmark}
\end{center}
\end{figure}

We present an example of landmark-driven matching in Fig.~\ref{fig:geodesic-landmark}. 
In this example, we study two BANs shown in the two left and the right corners of the top row. We use five landmarks -- one at a central node and two each placed automatically along the two longest paths in the graph
starting from the central node -- to perform improved registration of the graphs. The resulting geodesic is presented in the top row. We prune this display by removing the unmatched edges
and show the pruned geodesic in the bottom row. In this example, the geodesic distance
before using landmarks is $244.1698$ but comes down to $237.3467$ after using landmarks, 
signifying an improvement in the registration of nodes.  
 We demonstrate an example of landmark-based matching and mean computation in 
Fig.~\ref{fig:mean-landmark}. 
The use of landmarks improves the registration and thus keeps the major patterns (two longest paths) for the mean. 
 
\section{Conclusion}
This paper develops techniques for mathematically representing and statistically analyzing shapes of 
Brain Arterial Networks or BANs, broken into four major components. 
These objects -- BAN components -- are complicated due to arbitrary numbers, shapes, 
sizes, and connectivities of 3D arterial curves. We represent them using
elastic graphs and their adjacency matrices, where entries in adjacency matrices are the shapes of the corresponding edges (arteries). We solve for registration of nodes across graphs between compared
and use the shape space's geometry to compute geodesics, sample means, and PCA components of BANs. The sample means help capture prominent common characteristics of different 
BANs while covariance-PCA analysis helps represent individual variability in a tractable fashion. 

The subsequent data analysis shows that age has a significant effect on three components of 
BANs -- left, right, and bottom -- but does not affect the top component. We use the tools developed
here to visualize the nature of BAN shape variability resulting from aging. Visualizing
shapes along the principal directions or age-regression directions is an essential accomplishment of this framework. One can pinpoint arteries that undergo significant deformations and arteries that do not change much over time. 

The overall performance of this framework relies on the quality of registration of nodes across
graphs. We have shown using landmarks could potentially improve the registration quality. For a discussion on the current computation framework, we refer readers to the supplementary materials~[\cite{supplement}].

%

\begin{funding}
This research was supported in part by the grants NIH R01 MH120299, NSF DMS 1621787, 
and NSF DMS 1953087. 
\end{funding}

\begin{supplement}
\stitle{Supplement to "Statistical shape analysis of brain arterial networks (BAN)"}
\sdescription{Discussion on Computational Issues and Descriptive Statistics of BAN; GIFs of Fig \ref{fig:geo_simu} and Fig \ref{fig:brain_artery_geodesic}.}
\end{supplement}


\bibliographystyle{imsart-nameyear} 
\bibliography{egbib}       

\end{document}